\documentclass[10pt,journal,compsoc]{IEEEtran}
\ifCLASSOPTIONcompsoc
  \usepackage[nocompress]{cite}
\else
  \usepackage{cite}
\fi
\ifCLASSINFOpdf
  \usepackage[pdftex]{graphicx}
\else
  \usepackage[dvips]{graphicx}
\fi
\usepackage{amsmath}
\usepackage{url}
\usepackage{color}
\usepackage{hyperref}
\usepackage{bm}
\usepackage{nicefrac}
\usepackage{amssymb}
\usepackage{multirow}
\usepackage{multicol}
\usepackage{array}
\usepackage{amsthm}
\DeclareMathOperator*{\argmax}{arg\,max} 
\DeclareMathOperator*{\argmin}{arg\,min} 
\usepackage{algorithm, algpseudocode}
\usepackage{subfigure}
\usepackage{makecell}

\newtheorem{lemma}{Lemma}
\newtheorem{theorem}{Theorem}

\newtheorem{remark}{Remark}

\newcommand{\E}{\mathbb{E}}
\newcommand{\eat}[1]{}

\let\oldReturn\Return
\renewcommand{\Return}{\State\oldReturn}

\begin{document}
%
\title{Query-Efficient Black-box Adversarial Attacks Guided by a Transfer-based Prior}
%
%
%
%

\author{Yinpeng~Dong$^\dag$,
        Shuyu~Cheng$^\dag$,
        Tianyu~Pang,
        Hang~Su,
        and~Jun~Zhu$^\ddag$,~\IEEEmembership{Senior Member,~IEEE}
\IEEEcompsocitemizethanks{\IEEEcompsocthanksitem $^\dag$Y. Dong and S. Cheng contribute equally. 
\IEEEcompsocthanksitem $^\ddag$ J. Zhu is the corresponding author.
\IEEEcompsocthanksitem The authors are with the Dept. of Comp. Sci. and Tech., Institute for AI, Tsinghua-Bosch Joint ML Center, THBI Lab, BNRist Center, Tsinghua University, Beijing 100084, China; Pazhou Lab, Guangzhou, 510330, China.
        Email: \{dongyinpeng@mail, chengsy18@mails, pty17@mails, suhangss@mail, dcszj@mail\}.tsinghua.edu.cn}}

%
%

\markboth{Journal of \LaTeX\ Class Files,~Vol.~14, No.~8, August~2015}%
{Shell \MakeLowercase{\textit{et al.}}: Bare Demo of IEEEtran.cls for Computer Society Journals}
%



\IEEEtitleabstractindextext{%
\begin{abstract}
Adversarial attacks have been extensively studied in recent years since they can identify the vulnerability of deep learning models before deployed. In this paper, we consider the black-box adversarial setting, where the adversary needs to craft adversarial examples without access to the gradients of a target model. Previous methods attempted to approximate the true gradient either by using the transfer gradient of a surrogate white-box model or based on the feedback of model queries. However, the existing methods inevitably suffer from low attack success rates or poor query efficiency since it is difficult to estimate the gradient in a high-dimensional input space with limited information. To address these problems and improve black-box attacks, we propose two prior-guided random gradient-free (PRGF) 
algorithms based on biased sampling and gradient averaging, respectively. Our methods can take the advantage of a transfer-based prior given by the gradient of a surrogate model and the query information simultaneously. Through theoretical analyses, the transfer-based prior is appropriately integrated with model queries by an optimal coefficient in each method. Extensive experiments demonstrate that, in comparison with the alternative state-of-the-arts, both of our methods require much fewer queries to attack black-box models with higher success rates.
\end{abstract}

}

\maketitle

\IEEEdisplaynontitleabstractindextext

%
\IEEEpeerreviewmaketitle

\IEEEraisesectionheading{\section{Introduction}\label{sec:introduction}}

%
%
%
%

\IEEEPARstart{D}{espite} the significant success of deep learning models on various tasks~\cite{Goodfellow-et-al2016}, the security and reliability of these models have been challenged in the presence of adversarial examples~\cite{Biggio2013Evasion,Szegedy2013,Goodfellow2014,Kurakin2016}.
The maliciously crafted adversarial examples aim at causing misclassification of a target model by applying human imperceptible perturbations to natural examples.
It has garnered increasing attention to study the generation of adversarial examples (i.e., adversarial attack), which is indispensable to discover the weaknesses of deep learning algorithms~\cite{Szegedy2013,Athalye2018Obfuscated,dong2019evading}. Adversarial attacks therefore serve as a surrogate to evaluate robustness~\cite{carlini2016,madry2017towards,dong2020benchmarking}, and consequently contribute to the design of more robust deep learning models~\cite{Goodfellow2014,madry2017towards,pang2021tricks-at}.

Adversarial attacks are predominantly categorized into \emph{white-box} attacks and \emph{black-box} attacks according to different accessibility to the target model.
Getting access to the model architecture, parameters and especially gradients, an adversary can adopt various gradient-based methods~\cite{Goodfellow2014,Kurakin2016,carlini2016,madry2017towards} to generate adversarial examples under the white-box setting,
such as the fast gradient sign method (FGSM)~\cite{Goodfellow2014}, projected gradient descent method (PGD)~\cite{madry2017towards}, etc.
By contrast, under the more challenging black-box adversarial setting, the adversary has no or limited knowledge about the target model, and therefore needs to generate adversarial examples without any gradient information. In various real-world applications, the black-box setting is more practical than the white-box counterpart~\cite{ilyas2018black,Brendel2018Decision}.

Tremendous efforts have been made to develop black-box adversarial attacks~\cite{papernot2016practical,Chen2017ZOO,Brendel2018Decision,Dong2017,ilyas2018black,Bhagoji_2018_ECCV,tu2018autozoom,ilyas2018prior,dong2019efficient}.
A common idea of these techniques is to utilize an approximate gradient instead of the true but unknown gradient for generating adversarial examples. The approximate gradient can either stem from the gradient of a surrogate white-box model (termed as \textit{transfer-based} attacks) or be numerically estimated by the zeroth-order optimization algorithms (termed as \textit{query-based} attacks).

In transfer-based attacks, adversarial examples produced for a surrogate model are probable to remain adversarial for the target model due to the transferability~\cite{Papernot20162,Liu2016}. Recent methods have been introduced to improve the transferability by adopting a momentum optimizer~\cite{Dong2017} or performing input augmentations~\cite{xie2019improving,dong2019evading}. However, the success rate of transfer-based attacks is still far from satisfactory. 
This is because that there lacks an adjustment procedure when the gradient of the surrogate model points to a non-adversarial region of the target model.
In query-based attacks, the true gradient can be estimated by various methods, such as finite difference~\cite{Chen2017ZOO,Bhagoji_2018_ECCV}, random gradient estimation~\cite{tu2018autozoom} and natural evolution strategies~\cite{ilyas2018black}. Although these methods usually result in a higher attack success rate compared with the transfer-based attack methods~\cite{Chen2017ZOO,Bhagoji_2018_ECCV}, they inevitably require a tremendous number of queries to perform a successful attack.
The query inefficiency primarily comes from the under-utilization of priors, since the current methods are nearly optimal to estimate the gradient~\cite{ilyas2018prior}.


To overcome the aforementioned problems and improve black-box adversarial attacks, we propose two \textbf{prior-guided random gradient-free (PRGF)} algorithms based on biased sampling (BS) and gradient averaging (GA), respectively, which can utilize a transfer-based prior for query-efficient black-box attacks. 
The transfer-based prior originated from the gradient of a surrogate white-box model contains abundant prior knowledge of the true gradient.
Despite the same goal, the two proposed methods utilize the transfer gradient in different ways.
Specifically, our first method, abbreviated as PRGF-BS, provides a gradient estimate by querying the target model with random samples that are biased towards the transfer gradient and acquiring the corresponding loss values.
Our second method, denoted as PRGF-GA, performs a weighted average of the transfer gradient and the gradient estimate provided by the ordinary random gradient-free (RGF) method~\cite{nesterov2017random,ghadimi2013stochastic,duchi2015optimal}.
Under the gradient estimation framework, we provide theoretical analyses on deriving the optimal coefficients of controlling the strength of the transfer gradient in both algorithms.

Furthermore, our methods are flexible to integrate other prior information. As a concrete example, we incorporate the commonly used \emph{data-dependent prior}~\cite{ilyas2018prior} into our algorithms along with the transfer-based prior. We also provide theoretical analyses on how to embrace both priors appropriately.
Besides, we extend our methods to the scenario that multiple surrogate models are available, as studied in~\cite{Liu2016,Dong2017}, in which we can further boost the attack performance with a more effective transfer-based prior.
Extensive experiments demonstrate that  
both of our methods significantly outperform the previous state-of-the-art methods in terms of black-box attack success rate and query efficiency, verifying the superiority of our algorithms for black-box attacks.

This paper substantially extends and improves the conference version \cite{cheng2019improving}. We additionally propose a new PRGF algorithm based on gradient averaging and integrate it with the data-dependent prior. We also consider the scenario that multiple surrogate models are available and provide a subspace projection method to extract a more effective transfer-based prior. Besides, we conduct additional experiments by comparing more methods, using different surrogate models, and considering another dataset, to show the superiority of our methods. Overall, we make the following contributions:

\begin{itemize}
    \item[1)] We propose to improve black-box adversarial attacks by incorporating a transfer-based prior given by the gradient of a surrogate model. The transfer-based prior provides abundant prior information of the true gradient due to the adversarial transferability.
    \item[2)] We develop two prior-guided random gradient-free (PRGF) algorithms to utilize the transfer-based prior, based on biased sampling and gradient averaging, respectively. Theoretical analyses derive the optimal coefficients of integrating the transfer-based prior.
    \item[3)] We demonstrate the flexibility of our algorithms by incorporating the widely used data-dependent prior and considering multiple surrogate models.
    \item[4)] We validate that the proposed methods can improve the success rate of black-box adversarial attacks and reduce the requisite numbers of queries significantly compared with the state-of-the-art methods.
\end{itemize}

The rest of the paper is organized as follows. Section~\ref{sec:2} reviews the background and related work on black-box adversarial attacks. Section~\ref{sec:3} introduces the gradient estimation framework. Section~\ref{sec:4} and Section~\ref{sec:5} present the proposed PRGF algorithms, and their extensions with data-dependent priors and multiple surrogate models. 
Section~\ref{sec:6} presents empirical studies. Finally, Section~\ref{sec:7} concludes.

\section{Background}\label{sec:2}

\subsection{Adversarial Setup}
Given a classifier $C(x)$ and an input-label pair $(x, y)$ where $x\in\mathbb{R}^D$, the goal of attacks is to generate an adversarial example $x^{adv}$ that is misclassified by $C$ while the distance between the adversarial example $x^{adv}$ and the natural one $x$ measured by the $\ell_p$ norm is smaller than a threshold $\epsilon$ as 
\begin{equation}\label{eq:1}
   C(x^{adv}) \neq y, \text{ s.t. }  \|x^{adv}-x\|_p\leq\epsilon.
\end{equation}

Note that formulation~\eqref{eq:1} corresponds to an untargeted attack. We present our framework and algorithms based on untargeted attacks for clarity, while the extension to targeted ones is straightforward.

An adversarial example can be generated by solving the constrained optimization problem as 
\begin{equation}
    x^{adv} = \argmax_{x':\|x' - x\|_{p} \leq \epsilon} f(x', y),
    \label{eq:problem}
\end{equation}
where $f$ is a loss function on top of the classifier $C(x)$, e.g., the cross-entropy loss.
Several gradient-based methods~\cite{Goodfellow2014,Kurakin2016,carlini2016,madry2017towards,Dong2017} have been proposed to solve this optimization problem.
The typical projected gradient descent method (PGD)~\cite{madry2017towards} iteratively generates adversarial examples as 
\begin{equation}
x_{t+1}^{adv} = \Pi_{\mathcal{B}_p(x,\epsilon)} (x_t^{adv} + \eta\cdot g_t),
\label{eq:iter}
\end{equation}
where $\mathcal{B}_p(x,\epsilon)=\{x':\|x'-x\|_p\leq\epsilon\}$ denotes the $\ell_p$ ball centered at $x$ with radius $\epsilon$, $\Pi$ is the projection operation, $\eta$ is the step size, and $g_t$ is the normalized gradient under the $\ell_p$ norm, e.g., $g_t = \frac{\nabla_{x}f(x_t^{adv},y)}{\|\nabla_{x}f(x_t^{adv},y)\|_2}$ under the $\ell_2$ norm, and $g_t = \mathrm{sign}(\nabla_{x}f(x_t^{adv},y))$ under the $\ell_{\infty}$ norm.
Those methods such as PGD require full access to the gradients of the target model, which are known as white-box attacks.

\subsection{Black-box Attacks}\label{sec:2-2}
In contrast to white-box attacks, black-box attacks have no or limited knowledge about the target model, which can be challenging yet practical in various real-world applications.
We can still adopt the PGD method to generate adversarial examples, except that the true gradient $\nabla_{x}f(x,y)$ is usually replaced by an approximate gradient. 
Black-box attacks can be roughly divided into transfer-based attacks and query-based attacks. 
Transfer-based attacks depend on the gradient of a surrogate white-box model to generate adversarial examples, which are probable to fool the black-box model due to the transferability~\cite{Papernot20162,Liu2016}. 
Some query-based attacks estimate the gradient by the zeroth-order optimization methods, when the loss values could be accessed through queries. 
Chen et al.~\cite{Chen2017ZOO} propose to estimate the gradient at each coordinate by the symmetric difference quotient~\cite{lax2014calculus} as
\begin{equation}
\hat{g}_i = \frac{f(x+\sigma e_i,y)-f(x-\sigma e_i,y)}{2\sigma} \approx \frac{\partial f(x,y)}{\partial x_i},
\label{eq:finite-difference}
\end{equation}
where $\sigma$ is a small constant and $e_i$ is the $i$-th unit basis vector.
Although query-efficient mechanisms have been developed~\cite{Chen2017ZOO,Bhagoji_2018_ECCV}, the coordinate-wise gradient estimation inherently leads to the query complexity being proportional to the input dimension $D$, which is prohibitively large with a high-dimensional input space, e.g., $D\approx270$,$000$ for ImageNet~\cite{russakovsky2015imagenet}.
To improve query efficiency, the approximated gradient $\hat{g}$ can be obtained by the random gradient-free (RGF) method~\cite{nesterov2017random,ghadimi2013stochastic,duchi2015optimal} as
\begin{equation}
     \hat{g} = \frac{1}{q}\sum_{i=1}^{q}\hat{g}_i, \text{ with  } \hat{g}_i = \frac{f(x+\sigma u_i,y)-f(x,y)}{\sigma}\cdot u_i,
\label{eq:estimate}
\end{equation}
where $\{u_i\}_{i=1}^q$ are the random vectors sampled independently from a pre-defined distribution $\mathcal{P}$ on $\mathbb{R}^D$ and $\sigma$ is the parameter to control the sampling variance. It can be noted that $\hat{g}_i \rightarrow u_i^\top \nabla_x f(x,y) \cdot u_i$ when $\sigma\rightarrow 0$, which is nearly an unbiased estimator of the gradient when $\mathbb{E}[u_iu_i^\top] = \mathbf{I}$~\cite{duchi2015optimal}. In practice, the final gradient estimator $\hat{g}$ is averaged over $q$ random directions to reduce the variance.
\cite{ilyas2018black} relies on the natural evolution strategies (NES)~\cite{JMLR:v15:wierstra14a} to estimate the gradient, which is another variant of Eq.~\eqref{eq:estimate}. The difference is that \cite{ilyas2018black} conducts the antithetic sampling over a Gaussian distribution.
Ilyas et al.~\cite{ilyas2018prior} prove that these methods are nearly optimal to estimate the gradient, but the query efficiency could be improved by incorporating informative priors.
They identify the time- and data-dependent priors for black-box attacks. Different from those alternative methods, our adopted transfer-based prior is more effective as shown in the experiments. Moreover, the transfer-based prior can also be used simultaneously with other priors. We demonstrate the flexibility of our algorithms by incorporating the commonly used data-dependent prior as an example.

\subsection{Attacks based on both Transferability and Queries}
There are also several works that utilize both the transferability of adversarial examples and the model queries for black-box attacks.
A local substitute model can be trained to mimic the black-box model with a synthetic dataset, in which the labels are given by the black-box model through queries~\cite{papernot2016practical,Papernot20162}. Then the black-box model can be evaded by the adversarial examples crafted for the substitute model based on the transferability.
A meta-model~\cite{oh2017towards} can reverse-engineer the black-box model and predict its attributes (e.g.,  architecture, optimization procedure, and training samples) through a sequence of model queries.
Given the predicted attributes of the black-box model, the attacker can find similar surrogate models, which exhibit better transferability of the generated adversarial examples against the black-box model.
All of these methods use queries to obtain knowledge of the black-box model, and train/find surrogate models to generate adversarial examples, with the purpose of improving the transferability.
However, we do not optimize the surrogate model, but focus on utilizing the gradient(s) of a (multiple) fixed surrogate model(s) to obtain a more accurate gradient estimate.

Although a recent work~\cite{brunner2019guessing} also uses the gradient of a surrogate model to improve the query efficiency of black-box attacks, it focuses on a different attack scenario, where the adversary can only acquire the hard-label outputs, but we consider the adversarial setting that the loss values can be accessed. Moreover, this method controls the strength of the transfer gradient by a preset hyperparameter, but we obtain its optimal value through theoretical analyses based on the gradient estimation framework.
It is worth mentioning that a similar but independent work~\cite{maheswaranathan2018guided} also uses surrogate gradients to improve zeroth-order optimization, but they do not apply their method to black-box adversarial attacks.

\section{Gradient Estimation Framework}\label{sec:3}

Before we delve into the details of the proposed methods, we first introduce the gradient estimation framework in this section, which builds up the foundation of our theoretical analyses.

The key problem in black-box adversarial attacks is to estimate the gradient of a target model, which can then be used to carry out gradient-based attacks.
The goal of this work is to estimate the gradient $\nabla_{x}f(x,y)$ of the black-box model $f$
more accurately to improve black-box attacks.
We denote the gradient $\nabla_x f(x,y)$ by $\nabla f(x)$ in the sequel for notation clarity. We assume that $\nabla f(x)\neq 0$ in this paper.
The objective of gradient estimation is to find the best estimator that approximates the true gradient $\nabla f(x)$ by reaching the minimum value of the loss function as
\begin{equation}
    \hat{g}^* = \argmin_{\hat{g}\in \mathcal{G}} L(\hat{g}), 
\label{eq:ge_problem}
\end{equation}
where $\hat{g}$ is a gradient estimator given by any estimation algorithm, $\mathcal{G}$ is the set of all possible gradient estimators, and $L(\hat{g})$ is a loss function to evaluate the performance of the estimator $\hat{g}$.
Specifically, we let the loss function of the gradient estimator $\hat{g}$ be
\begin{equation}
    L(\hat{g}) = \min_{b\geq0} \E\|\nabla f(x)-b\hat{g}\|_2^2,
\label{eq:loss}
\end{equation}
where the expectation is taken over the randomness of the estimation algorithm to obtain $\hat{g}$. 
We define the loss $L(\hat{g})$ to be the minimum expected squared $\ell_2$ distance between the true gradient $\nabla f(x)$ and the scaled estimator $b\hat{g}$. 
The previous work~\cite{tu2018autozoom} considers the expected squared $\ell_2$ distance $\E\|\nabla f(x)-\hat{g}\|_2^2$ as the loss function, which is similar to ours. However, the value of their adopted loss function will change with different magnitudes of the estimator $\hat{g}$ (i.e., scaling $\hat{g}$ can cause varying loss values). 
In the process of generating adversarial examples, the gradient is usually normalized~\cite{Goodfellow2014,Kurakin2016,madry2017towards}, indicating that the direction of the gradient estimator, instead of the magnitude, will affect the performance of attacks. Thus, we incorporate a scaling factor $b$ in Eq.~\eqref{eq:loss} and minimize the error w.r.t. $b$, which can neglect the impact of the magnitude on the loss of the estimator $\hat{g}$.

\section{Methods}
\label{sec:4}
In this section, we present the two proposed \textbf{prior-guided random gradient-free (PRGF)} methods, which are variants of the ordinary random gradient-free (RGF) method. 
Recall that in RGF, the gradient is estimated through a set of random vectors $\{u_i\}_{i=1}^q$ as in Eq.~\eqref{eq:estimate} with $q$ being the total number.
Directly using RGF without prior information (i.e., sampling $u_i$ from an uninformative distribution such as a uniform distribution) will result in poor query efficiency as demonstrated in our experiments.
Therefore, we propose to improve the RGF estimator by utilizing the transfer-based prior, through either biased sampling or gradient averaging.

We denote the normalized transfer gradient of a surrogate model as $v$ such that $\|v\|_2=1$, and the cosine similarity between the transfer gradient and the true gradient as
\begin{equation}\label{eq:alpha}
    \alpha=v^\top\overline{\nabla f(x)}, \text{  with   } \overline{\nabla f(x)}=\frac{\nabla f(x)}{\|\nabla f(x)\|_2}, 
\end{equation}
where $\overline{\nabla f(x)}$ is the $\ell_2$ normalization of the true gradient $\nabla f(x)$.\footnote{ 
We use $\overline{e}$ to denote the $\ell_2$ normalization of a vector $e$ in this paper.}
We assume that $\alpha\geq 0$ without loss of generality, since we can reassign  $v\leftarrow -v$ when $\alpha<0$.

We will introduce the two PRGF methods in Section~\ref{sec:bs} and Section~\ref{sec:ga}, respectively.
As the true value of the cosine similarity $\alpha$ is unknown, we develop a method to estimate it efficiently, which will be introduced in Section~\ref{sec:alpha}.

\subsection{PRGF with Biased Sampling}\label{sec:bs}
Rather than sampling the random vectors $\{u_i\}_{i=1}^q$ from an uninformative distribution as the ordinary RGF method, our first proposed method samples the random vectors that are biased towards the transfer gradient $v$, to fully exploit the prior information.
For the gradient estimator $\hat{g}$ in Eq.~\eqref{eq:estimate}, we further assume that the sampling distribution $\mathcal{P}$ is defined on the unit hypersphere in the $D$-dimensional input space, such that the random vectors $\{u_i\}_{i=1}^q$ drawn from $\mathcal{P}$ satisfy $\|u_i\|_2=1$. Then, we can calculate the loss of the gradient estimator $\hat{g}$ in Eq.~\eqref{eq:estimate} by the following theorem.
\begin{theorem}\label{the:1}
(Proof in Appendix A.1) If $f$ is differentiable at $x$, the loss of the gradient estimator $\hat{g}$ defined in Eq.~\eqref{eq:estimate} is
\begin{equation}
\begin{gathered}
    \lim_{\sigma\to 0}L(\hat{g})=\|\nabla f(x)\|_2^2 \hspace{30ex}\\-\frac{\big(\nabla f(x)^\top\mathbf{C}\nabla f(x)\big)^2}{(1-\frac{1}{q})\nabla f(x)^\top\mathbf{C}^2\nabla f(x)+\frac{1}{q}\nabla f(x)^\top\mathbf{C}\nabla f(x)},
    \label{eq:biased_obj}
\end{gathered}
\end{equation}
where $\sigma$ is the sampling variance, $\mathbf{C}=\E[u_iu_i^\top]$ with $u_i$ being the random vector, $\|u_i\|_2=1$, and $q$ is the number of random vectors as in Eq.~\eqref{eq:estimate}.  
\end{theorem}

It can be noted from Theorem~\ref{the:1} that we can minimize $L(\hat{g})$ by optimizing $\mathbf{C}$, i.e., we can obtain an optimal gradient estimator by appropriately sampling the random vectors $u_i$, yielding an query-efficient adversarial attack.
Given the definition of $\mathbf{C}$, it needs to satisfy two constraints: (1) it should be positive semi-definite; (2) its trace should be $1$ since $\mathrm{Tr}(\mathbf{C})=\E[\mathrm{Tr}(u_i u_i^\top)]=\E[u_i^\top u_i]=1$.

Specifically, $\mathbf{C}$ can be decomposed as $\sum_{j=1}^D \lambda_j v_j v_j^\top$, in which $\{\lambda_j\}_{j=1}^D$ and $\{v_j\}_{j=1}^D$ are the non-negative eigenvalues and the orthonormal eigenvectors of $\mathbf{C}$, satisfying $\sum_{j=1}^D \lambda_j=1$.
In our method, we propose to sample $u_i$ that are biased towards the transfer gradient $v$ to exploit its prior information. So we specify an eigenvector of $\mathbf{C}$ to be $v$, and let the corresponding eigenvalue be a tunable coefficient.
For the other eigenvalues, we set them to be equal since we do not have any prior knowledge about the other eigenvectors.
To this end, we let
\begin{equation}
    \mathbf{C}=\lambda v v^\top + \frac{1-\lambda}{D-1}(\mathbf{I}-v v^\top),
    \label{eq:mix_C}
\end{equation}
where $\lambda \in [0,1]$ controls the strength of the transfer gradient that the random vectors $\{u_i\}_{i=1}^q$ are biased towards.
We can easily construct a random vector with unit length while satisfying Eq.~\eqref{eq:mix_C} as (proof in Appendix A.2)
\begin{equation}
    u_i=\sqrt{\lambda}\cdot v+\sqrt{1-\lambda}\cdot\overline{(\mathbf{I}-vv^\top)\xi_i},
    \label{eq:sample_rv}
\end{equation}
where $\xi_i$ is sampled uniformly from the $D$-dimensional unit hypersphere. Hereby, the problem becomes optimizing $\lambda$ that minimizes $L(\hat{g})$. 
Note that when $\lambda=\frac{1}{D}$ and  $\mathbf{C}=\frac{1}{D}\mathbf{I}$, such that the random vectors are drawn from the uniform distribution on the hypersphere, our method degenerates into the ordinary RGF method.
When $\lambda\in[0,\frac{1}{D})$, it indicates that the transfer gradient is worse than a random vector, so we are encouraged to search in other directions by using a small $\lambda$.

To find the optimal $\lambda$ that leads to the minimum value of the loss $L(\hat{g})$, we plug Eq.~\eqref{eq:mix_C} into Eq.~\eqref{eq:biased_obj}, and obtain the closed-form solution as (proof in Appendix A.3)
\begin{align}\small
    \lambda^* =
    \begin{cases}
        \hfil 0 & \text{if } \alpha^2\in[0,a_l]  \\
        \dfrac{(1-\alpha^2)(\alpha^2(D+2q-2)-1)}{2\alpha^2 Dq - \alpha^4 D(D+2q-2) - 1} & \text{if } \alpha^2 \in (a_l, a_r) \\
        \hfil 1 & \text{if } \alpha^2 \in [a_r, 1]
    \end{cases}
    \label{eq:lambda-1}
\end{align}
where $a_l=\frac{1}{D+2q-2}$ and $a_r=\frac{2q-1}{D+2q-2}$ (recall that $\alpha$ is the cosine similarity defined in Eq.~\eqref{eq:alpha}).

\begin{remark}\label{remark:1}
It can be proven (in Appendix A.4) that $\lambda^*$ is a monotonically increasing function of $\alpha^2$, and a monotonically decreasing function of $q$ (when $\alpha^2>\frac{1}{D}$).
It indicates that a larger $\alpha$ or a smaller $q$ (when the transfer gradient is not worse than a random vector) would result in a larger $\lambda^*$, which makes sense since we tend to rely on the transfer gradient more when (1) it approximates the true gradient better; (2) the number of queries is not enough to provide much gradient information.
\end{remark}

\begin{algorithm}[t]
\small
\caption{Prior-guided random gradient-free algorithm based on biased sampling (PRGF-BS)}
\label{alg:biased}
\begin{algorithmic}[1]
\Require The black-box model $f$; input $x$ and label $y$; the normalized transfer gradient $v$; sampling variance $\sigma$; number of queries $q$; input dimension $D$.
\Ensure Estimate of the gradient $\nabla f(x)$.
\State Estimate the cosine similarity $\alpha=v^\top\overline{\nabla f(x)}$ (detailed in Section~\ref{sec:alpha});
\State Calculate $\lambda^*$ according to Eq.~\eqref{eq:lambda-1} given $\alpha$, $q$, and $D$;
\If {$\lambda^*=1$}
\Return $v$;
\EndIf
\State $\hat{g} \leftarrow \mathbf{0}$;
\For {$i = 1$ to $q$}
\State Sample $\xi_i$ from the uniform distribution on the $D$-dimensional unit hypersphere;
\State $u_i=\sqrt{\lambda^*}\cdot v+\sqrt{1-\lambda^*}\cdot\overline{(\mathbf{I}-vv^\top)\xi_i}$;

\State $\hat{g} \leftarrow \hat{g} + \dfrac{f(x + \sigma u_i,y) - f(x,y)}{\sigma} \cdot u_i$;
\EndFor
\Return $\nabla f(x)\leftarrow \dfrac{1}{q}\hat{g}$.
\end{algorithmic}
\end{algorithm}

We summarize the PRGF-BS algorithm in Algorithm~\ref{alg:biased}.
Note that when $\lambda^*=1$, we do not need to sample $q$ random vectors because they all equal to $v$, and we directly return the transfer gradient $v$ as the estimate of $\nabla f(x)$ (Step 3-5), which can save many queries.


\subsection{PRGF with Gradient Averaging}\label{sec:ga}

In this section, we propose an alternative method to incorporate the transfer gradient $v$ based on gradient averaging.
The motivation is as follows.
We observe that the RGF estimator in Eq.~\eqref{eq:estimate} has the form $\hat{g}=\frac{1}{q}\sum_{i=1}^q\hat{g}_i$, where multiple rough estimates are averaged. Indeed, the transfer gradient itself can also be considered as an estimate of the true gradient. Thus it is reasonable to perform a weighted average of the transfer gradient and the RGF estimator. 

In particular, we first obtain the ordinary RGF estimator defined in Eq.~\eqref{eq:estimate} with the sampling distribution $\mathcal{P}$ being the uniform distribution on the $D$-dimensional unit hypersphere, which is denoted as $\hat{g}^U$. Then we normalize $\hat{g}^U$ and perform a weighted average of the normalized transfer gradient $v$ and the normalized RGF estimator $\overline{\hat{g}^U}$ as
\begin{equation}
    \hat{g} = \mu v+(1-\mu) \overline{\hat{g}^U},
    \label{eq:average}
\end{equation}
where $\mu\in [0,1]$ is a balancing coefficient playing a similar role as $\lambda$ in PRGF-BS. 

Given the gradient estimator in Eq.~\eqref{eq:average}, we also aim at deriving the optimal $\mu$ that minimizes the loss of the estimator $L(\hat{g})$. We let $\beta=\overline{\frac{1}{q}\sum_{i=1}^q (u_i^\top \nabla f(x)\cdot u_i)}^\top\overline{\nabla f(x)}$ be the cosine similarity between $\frac{1}{q}\sum_{i=1}^q (u_i^\top \nabla f(x)\cdot u_i)$ and the true gradient $\nabla f(x)$, where $\{u_i\}_{i=1}^q$ are sampled from the uniform distribution. 
As discussed in Section~\ref{sec:2-2}, the RGF estimator $\hat{g}^U\rightarrow\frac{1}{q}\sum_{i=1}^q (u_i^\top \nabla f(x)\cdot u_i)$ when $\sigma\rightarrow0$, and consequently $\beta\rightarrow \overline{\hat{g}^U}^\top\overline{\nabla f(x)}$ as the cosine similarity between the ordinary RGF estimator and the true gradient.
Recall that $\alpha = v^\top\overline{\nabla f(x)}$ is the cosine similarity between the transfer gradient and the true gradient. Then we have the following theorem on the loss of the gradient estimator in Eq.~\eqref{eq:average}.

\begin{theorem}\label{thm:average}
    (Proof in Appendix A.5) If $f$ is differentiable at $x$, the loss of the gradient estimator defined in Eq.~\eqref{eq:average} is
    \begin{equation}
    \begin{gathered}
        \lim_{\sigma\to 0} L(\hat{g})=\|\nabla f(x)\|_2^2 \hspace{28ex} \\ -\frac{(\mu\alpha+(1-\mu)\E[\beta])^2}{\mu^2+(1-\mu)^2+2\mu(1-\mu)\alpha\E[\beta]}\|\nabla f(x)\|_2^2,
        \label{eq:theorem-2}
    \end{gathered}
    \end{equation}
    where $\sigma$ is the sampling variance to get $\hat{g}^U$.
\end{theorem}

Theorem~\ref{thm:average} indicates that we can achieve the minimum value of $ L(\hat{g})$ by optimizing $\mu$. We can calculate the closed-form solution of the optimal $\mu$ as (proof in Appendix A.6)
\begin{equation}
    \mu^*=\frac{\alpha(1-\E[\beta]^2)}{\alpha(1-\E[\beta]^2) + (1-\alpha^2)\E[\beta]}.
    \label{eq:mu-1}
\end{equation}
\begin{remark}\label{remark:2}
We can easily see that $\mu^*$ is a monotonically increasing function of $\alpha$, as well as a monotonically decreasing function of $\E[\beta]$. As will shown in Eq.~\eqref{eq:ebeta}, a larger number of queries $q$ for the RGF estimator can result in a larger $\E[\beta]$, such that $\mu^*$ is a monotonically decreasing function of $q$. These conclusions are consistent with our intuition as explained in Remark~\ref{remark:1}.
\end{remark}

Although we have derived the optimal $\mu$ in Eq.~\eqref{eq:mu-1}, the true value of $\E[\beta]$ is still unknown. We find that $\E[\beta]$ cannot directly be calculated but can roughly be approximated as (proof in Appendix A.7)
\begin{equation}\label{eq:ebeta}
    \E[\beta] \approx \sqrt{\frac{q}{D+q-1}},
\end{equation}
where $D$ and $q$ are the input dimension and the number of queries to get $\hat{g}^U$, respectively. Note that $\E[\beta]$ is irrelevant to the true gradient $\nabla f(x)$. We find such an approximation works well in practice.

\begin{algorithm}[t]
\small
\caption{Prior-guided random gradient-free algorithm based on gradient averaging (PRGF-GA)}
\label{alg:average}
\begin{algorithmic}[1]
\Require The black-box model $f$; input $x$ and label $y$; the normalized transfer gradient $v$; sampling variance $\sigma$; number of queries $q$; input dimension $D$; threshold $c$.
\Ensure Estimate of the gradient $\nabla f(x)$.
\State Estimate the cosine similarity $\alpha=v^\top\overline{\nabla f(x)}$ (detailed in Section~\ref{sec:alpha});
\State Approximate $\E[\beta]$ by $\sqrt{\frac{q}{D+q-1}}$ as in Eq.~\eqref{eq:ebeta};
\State Calculate $\mu^*$ according to Eq.~\eqref{eq:mu-1} given $\alpha$ and $\E[\beta]$;
\If {$\mu^*\geq c$}
\Return $v$;
\EndIf
\State $\hat{g}^U \leftarrow \mathbf{0}$;
\For {$i = 1$ to $q$}
\State Sample $u_i$ from the uniform distribution on the $D$-dimensional unit hypersphere;
\State $\hat{g}^U \leftarrow \hat{g}^U + \dfrac{f(x + \sigma u_i,y) - f(x,y)}{\sigma} \cdot u_i$;
\EndFor
\Return $\nabla f(x)\leftarrow \mu^* v+(1-\mu^*)\overline{\hat{g}^U}$.
\end{algorithmic}
\end{algorithm}

It should be noted that we have $\mu^*<1$, which means that we always need to take $q$ queries to get $\hat{g}^U$. However, when $\mu^*$ is close to $1$, the improvement of using $\hat{g}=\mu^* v+(1-\mu^*)\overline{\hat{g}^U}$ instead of directly using $v$ as the estimate is marginal. 
But the former requires $q$ more queries than the latter. 
To save queries, we use the transfer gradient $v$ as the estimate of $\nabla f(x)$ when it approximates $\nabla f(x)$ well. Thus we preset a threshold $c\in (0,1)$ such that when $\mu^*\geq c$, we return $v$ directly as the gradient estimate.
We summarize the overall PRGF-GA algorithm in Algorithm~\ref{alg:average}.\footnote{The actual implementation of PRGF-GA is slightly different from Algorithm~\ref{alg:average}, which will be explained in Appendix B.}

\begin{figure}[h]
\centering
\vspace{-1ex}
\includegraphics[width=0.9\columnwidth]{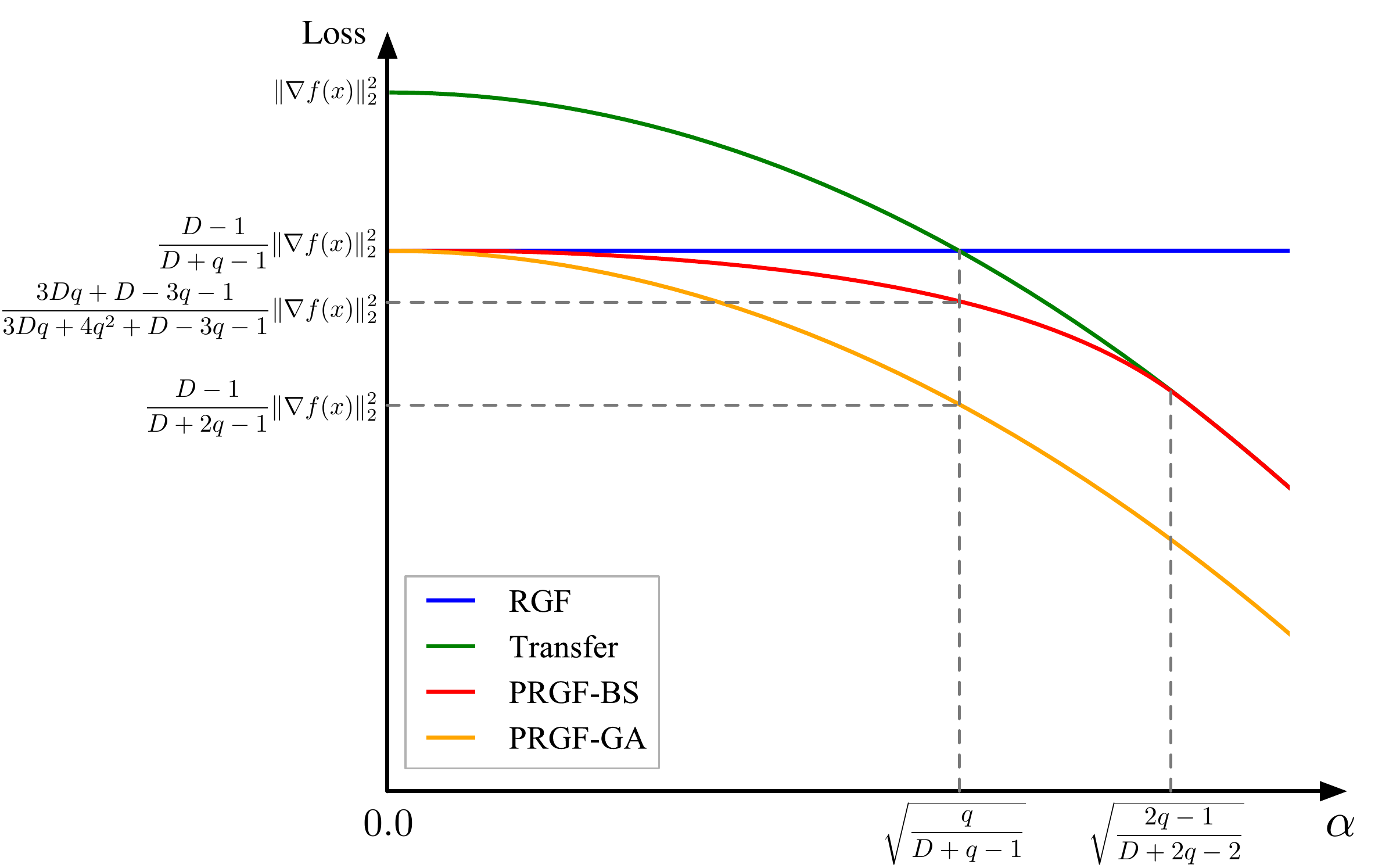}
\vspace{-2ex}
\caption{The loss curves of the different gradient estimators w.r.t. $\alpha$. The loss of the RGF estimator is $\frac{D-1}{D+q-1}\|\nabla f(x)\|_2^2$. The loss of the transfer gradient is $(1-\alpha^2)\|\nabla f(x)\|_2^2$. The loss of the PRGF-BS and PRGF-GA estimators can be derived by plugging $\lambda^*$ and $\mu^*$ into Eq.~\eqref{eq:biased_obj} and Eq.~\eqref{eq:theorem-2}, respectively.}
\label{fig:theory}
\vspace{-2ex}
\end{figure}

\vspace{2ex}
\noindent \textbf{Comparisons between PRGF-BS and PRGF-GA.} Because the two proposed methods utilize the transfer-based prior in different ways, we are interested in the loss (in Eq.~\eqref{eq:loss}) of the gradient estimators given by different methods, as well as the improvements over the ordinary RGF estimator and the transfer-based prior. To this end, we show the loss curves of gradient estimators given by RGF, transfer gradient, PRGF-BS, and PRGF-GA, respectively, w.r.t. different $\alpha$, in Fig.~\ref{fig:theory}. PRGF-GA can get a lower loss value than PRGF-BS with a given $\alpha$, indicating that PRGF-GA can utilize the transfer-based prior better. This is also verified in the experiments.

\subsection{Estimation of Cosine Similarity}
\label{sec:alpha}
To complete our algorithms, we need to estimate the cosine similarity $\alpha=v^\top\overline{\nabla f(x)} =\frac{v^\top\nabla f(x)}{\|\nabla f(x)\|_2}$, where $v$ is the normalized transfer gradient.
Note that the inner product $v^\top\nabla f(x)$ can directly be estimated by the finite difference method as
\begin{equation}
    v^\top\nabla f(x) \approx \frac{f(x+\sigma v, y)-f(x,y)}{\sigma}, 
    \label{eq:inner-product}
\end{equation}
with a small $\sigma$. Hence, the problem is reduced to estimating the norm of the gradient $\|\nabla f(x)\|_2$.


The basic method of estimating $\|\nabla f(x)\|_2$ is to adopt a $r$-degree homogeneous function $g$ of $S$ variables, i.e., $g(az)=a^r g(z)$ where $a\in\mathbb{R}$ and $z\in \mathbb{R}^S$. Then we have
\begin{align}
\label{norm_est}
    g\big(\mathbf{W}^\top\nabla f(x)\big)=\|\nabla f(x)\|_2^r \cdot g\big(\mathbf{W}^\top\overline{\nabla f(x)}\big), 
\end{align}
where $\mathbf{W}=[w_1, ..., w_S]$ denotes the matrix consisting of the $S$ random vectors $\{w_s\}_{s=1}^S$. Based on Eq.~\eqref{norm_est}, the norm of the gradient $\|\nabla f(x)\|_2$ could be computed easily if both $g\big(\mathbf{W}^\top\nabla f(x)\big)$ and $g\big(\mathbf{W}^\top\overline{\nabla f(x)}\big)$ can be obtained.

Suppose that we utilize $S$ queries to estimate $\|\nabla f(x)\|_2$. We draw a set of $S$ random vectors $\{w_s\}_{s=1}^S$ independently and uniformly from the $D$-dimensional unit hypersphere, and then estimate $w_s^\top\nabla f(x)$ based on Eq.~\eqref{eq:inner-product}. 
Given the estimated $w_s^\top\nabla f(x)$, we can obtain $g\big(\mathbf{W}^\top\nabla f(x)\big)$ directly.

However, it is non-trivial to obtain the value of $w_s^\top\overline{\nabla f(x)}$ as well as the function value $g\big(\mathbf{W}^\top\overline{\nabla f(x)}\big)$.
Nevertheless, we note that the distribution of $w_s^\top\overline{\nabla f(x)}$ is the same regardless of the direction of $\overline{\nabla f(x)}$, thus we can compute the expectation of the function value $\E\big[g\big(\mathbf{W}^\top\overline{\nabla f(x)}\big)\big]$.
Based on that, we use $\frac{g(\mathbf{W}^\top\nabla f(x))}{\E[g(\mathbf{W}^\top\overline{\nabla f(x)})]}$ as an unbiased estimator of $\|\nabla f(x)\|_2^r$. In particular, we choose $g$ as $g(z)=\frac{1}{S}\sum_{s=1}^S z_s^2$. Then $r=2$, and we have 
\begin{equation}
\begin{split}
\label{eq:g}
    \E\big[g\big(\mathbf{W}^\top\overline{\nabla f(x)}\big)\big]& = \E \big[(w_1^\top\overline{\nabla f(x)})^2] \\ &=\overline{\nabla f(x)}^\top \E[w_1w_1^\top]\overline{\nabla f(x)}=\frac{1}{D}.
\end{split}
\end{equation}
By plugging Eq.~\eqref{eq:g} into Eq.~\eqref{norm_est}, we can obtain the estimate of the gradient norm as
\begin{equation}
\begin{split}
    \|\nabla f(x)\|_2  \approx& \sqrt{\frac{D}{S}\sum_{s=1}^S \left(\frac{f(x+\sigma w_s,y) - f(x,y)}{\sigma}\right)^2}.
    \end{split}
\end{equation}


To save queries, we estimate the gradient norm periodically instead of in every iteration, since usually it does not change very fast in the optimization process.

\section{Extensions}\label{sec:5}
In this section, we extend our algorithms for incorporating the data-dependent prior and adopting multiple surrogate models to give the transfer-based prior.

\subsection{Data-dependent Prior}
\label{sec:dp}

The commonly used data-dependent prior~\cite{ilyas2018prior} is proposed to reduce the query complexity, which suggests that we can utilize the structure of the inputs to reduce the input space dimension without sacrificing much accuracy of gradient estimation. 
The idea of reducing the input dimension has already been adopted in several 
works~\cite{Chen2017ZOO,tu2018autozoom,guo2018low,brunner2019guessing}, which has shown promise for query-efficient black-box attacks.
We observe that many works restrict the adversarial perturbations to lie in a linear subspace of the input space, which allows the application of our theoretical framework.
Specifically, we focus on the data-dependent prior proposed in~\cite{ilyas2018prior}. Below we introduce how to incorporate it into RGF, PRGF-BS, and PRGF-GA appropriately.

\textbf{RGF.} For the RGF gradient estimator in Eq.~\eqref{eq:estimate},
to leverage the data-dependent prior, suppose that $u_i=\mathbf{V}\xi_i$, where $\mathbf{V}=[v_1,v_2,...,v_d]$ is a $D\times d$ matrix ($d<D$), $\{v_j\}_{j=1}^d$ is an orthonormal basis in the $d$-dimensional subspace of the input space, and $\xi_i$ is a random vector sampled from the $d$-dimensional unit hypersphere. In~\cite{ilyas2018prior}, the random vector $\xi_i$ drawn in $\mathbb{R}^d$ is up-sampled to $u_i$ in $\mathbb{R}^D$ by the nearest neighbor algorithm. The orthonormal basis $\{v_j\}_{j=1}^d$ can be obtained by first up-sampling the standard basis in $\mathbb{R}^d$ with the same method and then applying normalization.
For the ordinary RGF method, $\xi_i$ is sampled uniformly from the $d$-dimensional unit hypersphere, and $\mathbf{C}=\frac{1}{d}\sum_{j=1}^d v_j v_j^\top$ (recall that $\mathbf{C}=\E[u_iu_i^\top]$ as defined in Theorem~\ref{the:1}).

\textbf{PRGF-BS.} For PRGF with biased sampling, we consider incorporating the data-dependent prior into the algorithm along with the transfer-based prior.
Similar to Eq.~\eqref{eq:mix_C}, we let one eigenvector of $\mathbf{C}$ be $v$ to exploit the transfer-based prior, and the others are given by the orthonormal basis in the subspace to exploit the data-dependent prior, as
\begin{equation}
    \mathbf{C}=\lambda v v^\top + \frac{1-\lambda}{d}\sum_{j=1}^d v_j v_j^\top.
    \label{eq:mix_dp_C}
\end{equation}
By plugging Eq.~\eqref{eq:mix_dp_C} into Eq.~\eqref{eq:biased_obj}, we can similarly obtain the optimal $\lambda$ as (proof in Appendix A.8)
\begin{align}
\small
    \hspace{-0.7ex}\lambda^* =
    \begin{cases}
        \hfil 0 & \text{if} \; \alpha^2 \in [0, a_l] \\
        \dfrac{A^2(A^2-\alpha^2(d+2q-2))}{A^4+\alpha^4d^2-2A^2\alpha^2(q+dq-1)} & \text{if} \; \alpha^2\in(a_l,a_r)\\
        \hfil 1 & \text{if} \; \alpha^2 \in [a_r, 1]
    \end{cases}
    \label{eq:lambda-2}
\end{align}
where $A^2=\sum_{j=1}^d (v_j^\top\overline{\nabla f(x)})^2$, $a_l = \frac{A^2}{d+2q-2}$, and $a_r = \frac{A^2(2q-1)}{d}$. Note that $A$ is unknown, which should also be estimated. We use a method similar to the one for estimating $\alpha$, which is detailed in Appendix C.

The remaining problem is to construct a random vector $u_i$ satisfying $\E[u_i u_i^\top]=\mathbf{C}$, with $\mathbf{C}$ specified in Eq.~\eqref{eq:mix_dp_C}. In general, this is difficult since $v$ is not orthogonal to the subspace.
To address this problem, we sample $u_i$ in a way that $\E[u_i u_i^\top]$ is a good approximation of $\mathbf{C}$ (explanation in Appendix A.9), which is similar to Eq.~\eqref{eq:sample_rv} as
\begin{equation}
    u_i=\sqrt{\lambda}\cdot v+\sqrt{1-\lambda}\cdot\overline{(\mathbf{I}-vv^\top)\mathbf{V}\xi_i},
    \label{eq:sampling-dp}
\end{equation}
where $\xi_i$ is sampled uniformly from the $d$-dimensional unit hypersphere. 

The PRGF-BS algorithm with the data-dependent prior is similar to Algorithm~\ref{alg:biased}.
We first estimate $\alpha$ and $A$, and then calculate $\lambda^*$ by Eq.~\eqref{eq:lambda-2}.
If $\lambda^*=1$, we use the transfer gradient $v$ as the estimate.
Otherwise, we sample $q$ random vectors by Eq.~\eqref{eq:sampling-dp} and get the gradient estimate by Eq.~\eqref{eq:estimate}.

\textbf{PRGF-GA.} We similarly incorporate the data-dependent prior into the PRGF-GA algorithm. 
In this case, we first get an ordinary subspace RGF estimator $\hat{g}^S$ instead of the ordinary RGF estimator, by sampling $\xi_i$ uniformly from the $d$-dimensional unit hypersphere and letting $u_i=\mathbf{V}\xi_i$. Then we normalize $\hat{g}^S$ and obtain the averaged gradient estimator in a similar manner to Eq.~\eqref{eq:average} as
\begin{equation}
    \hat{g} = \mu v+(1-\mu)\overline{\hat{g}^S}.
    \label{eq:dp_average}
\end{equation}

To derive the optimal $\mu$ that minimizes the loss $L(\hat{g} )$, we define $\overline{\nabla f(x)}_T=(\sum_{j=1}^d v_j v_j^\top)\overline{\nabla f(x)}$ as the projection of $\overline{\nabla f(x)}$ onto the subspace corresponding to the data-dependent prior. We also need $A^2=\sum_{j=1}^d (v_j^\top \overline{\nabla f(x)})^2=\|\overline{\nabla f(x)}_T\|^2$.
We let $\beta=\overline{\frac{1}{q}\sum_{i=1}^q (u_i^\top \nabla f(x)\cdot u_i)}^\top\overline{\nabla f(x)}$ be the cosine similarity between $\frac{1}{q}\sum_{i=1}^q (u_i^\top \nabla f(x)\cdot u_i)$ and the true gradient $\nabla f(x)$, in which $\{u_i\}_{i=1}^q$ lie in the subspace. 
We have the following theorem on the loss of the gradient estimator in Eq.~\eqref{eq:dp_average}.

\begin{theorem}\label{the:3}
(Proof in Appendix A.10) Let $\alpha_1=v^\top\overline{\nabla f(x)}_T$. If $f$ is differentiable at $x$ and $A^2>0$, the loss of the gradient estimator define in Eq.~\eqref{eq:dp_average} is
\begin{equation}
\begin{gathered}
    \lim_{\sigma\to 0} L(\hat{g})=\|\nabla f(x)\|^2 \hspace{28ex}\\ -\frac{(\mu\alpha+(1-\mu)\E[\beta])^2}{\mu^2+(1-\mu)^2+2\mu(1-\mu)\frac{\alpha_1}{A^2}\E[\beta]}\|\nabla f(x)\|^2,
    \label{eq:theorem-3}
\end{gathered}
\end{equation}
where $\sigma$ is the sampling variance to get $\hat{g}^S$.
\end{theorem}

Based on Theorem~\ref{the:3}, we calculate the optimal solution of $\mu$ by minimizing Eq.~\eqref{eq:theorem-3} as (proof in Appendix A.11)
\begin{equation}
    \mu^*=\frac{A^2\alpha-\alpha_1\E[\beta]^2}{(A^2-\alpha_1\E[\beta])(\alpha+\E[\beta])}\approx \frac{\alpha}{\alpha + \E[\beta]}.
    \label{eq:mu-2}
\end{equation}
The approximation works mainly because $A\gg \E[\beta]$ (since $\E[\beta]\approx A\sqrt{\frac{q}{d+q-1}}$ as shown Appendix A.11). Therefore, $\mu^*$ can be approximated without $\alpha_1$, such that we do not need to estimate $\alpha_1$.

The PRGF-GA algorithm with the data-dependent prior is similar to Algorithm~\ref{alg:average}.
We first estimate $\alpha$ and $A$, approximate $\E[\beta]$ by $A\sqrt{\frac{q}{d+q-1}}$, and then calculate $\mu^*$ by Eq.~\eqref{eq:mu-2}.
If $\mu^*\geq c$, we use the transfer gradient $v$ as the estimate.
Otherwise, we get the ordinary subspace RGF estimate $\hat{g}^S$ with $q$ queries, and then use $\hat{g}\leftarrow \mu^* v+(1-\mu^*)\overline{\hat{g}^S}$.

\subsection{Multiple Surrogate Models}
\label{sec:5-2}

The idea of utilizing multiple surrogate models has been adopted in~\cite{Liu2016,Dong2017} for improving transfer-based black-box attacks. They show that the adversarial examples generated for multiple models are more likely to fool other black-box models with the increased transferability. In our algorithms, we can also utilize multiple surrogate models to extract a more effective transfer-based prior, which can consequently enhance the attack performance.

Assume that we have $M$ surrogate models. For an input $x$, we denote the gradients of these surrogate models at $x$ as $\{g^{(m)}\}_{m=1}^M$, where the gradients are not normalized for now.
A simple approach to obtain the transfer-based prior is averaging these gradients directly, as $v=\overline{\frac{1}{M}\sum_{m=1}^M g^{(m)}}$.
Despite the simplicity, this approach treats the gradients of surrogate models with equal importance and neglects the intrinsic similarity between different surrogate models and the target model. It has been observed that the adversarial examples are more likely to transfer within the same family of model architectures~\cite{su2018robustness}, indicating that we could design an improved transfer-based prior by leveraging more useful surrogate models/gradients.

Specifically, we denote the $M$-dimensional subspace spanned by $\{g^{(m)}\}_{m=1}^M$ as $\mathbf{G}$. The best approximation of the true gradient $\nabla f(x)$ that lies in $\mathbf{G}$ is the projection of $\nabla f(x)$ onto the subspace $\mathbf{G}$. Therefore, we first get an orthonormal basis of $\mathbf{G}$ by the Gram–Schmidt orthonormalization method, denoted as $\{v^{(m)}\}_{m=1}^M$. Then the projection of $\nabla f(x)$ onto $\mathbf{G}$ can be expressed as
\begin{equation}
    \nabla f(x)_{\mathbf{G}} = \sum_{m=1}^M \nabla f(x)^\top v^{(m)}\cdot v^{(m)},
\end{equation}
in which the inner product $\nabla f(x)^\top v^{(m)}$ can be approximated by the finite difference method as shown in Eq.~\eqref{eq:inner-product}. Hence, we let the transfer-based prior be $v=\overline{\nabla f(x)_{\mathbf{G}}}$. With $v$ obtained by multiple surrogate gradients, we then perform PRGF-BS or PRGF-GA attacks with the same algorithms.

\section{Experiments}\label{sec:6}
In this section, we present the empirical results to demonstrate the effectiveness of the proposed methods on attacking black-box image classifiers. We perform untargeted attacks under both the $\ell_2$ and $\ell_\infty$ norms on the ImageNet~\cite{russakovsky2015imagenet} and CIFAR-10~\cite{krizhevsky2009learning} datasets. We show the results under the $\ell_2$ norm in this section and leave the extra results under the $\ell_\infty$ norm in Appendix D. The results for both norms are consistent to verify the superiority of our methods.
We also conduct experiments on defense models in Appendix E.
We first specify the experimental setting in Section~\ref{sec:6-0}. Then we show the performance of gradient estimation in  Section~\ref{sec:6-1}.
We further compare the attack performance of the proposed algorithms with others on ImageNet in Section~\ref{sec:6-2}, and on CIFAR-10 in Section~\ref{sec:6-3}, respectively.

\begin{figure*}[t]
\centering
\includegraphics[width=0.94\linewidth]{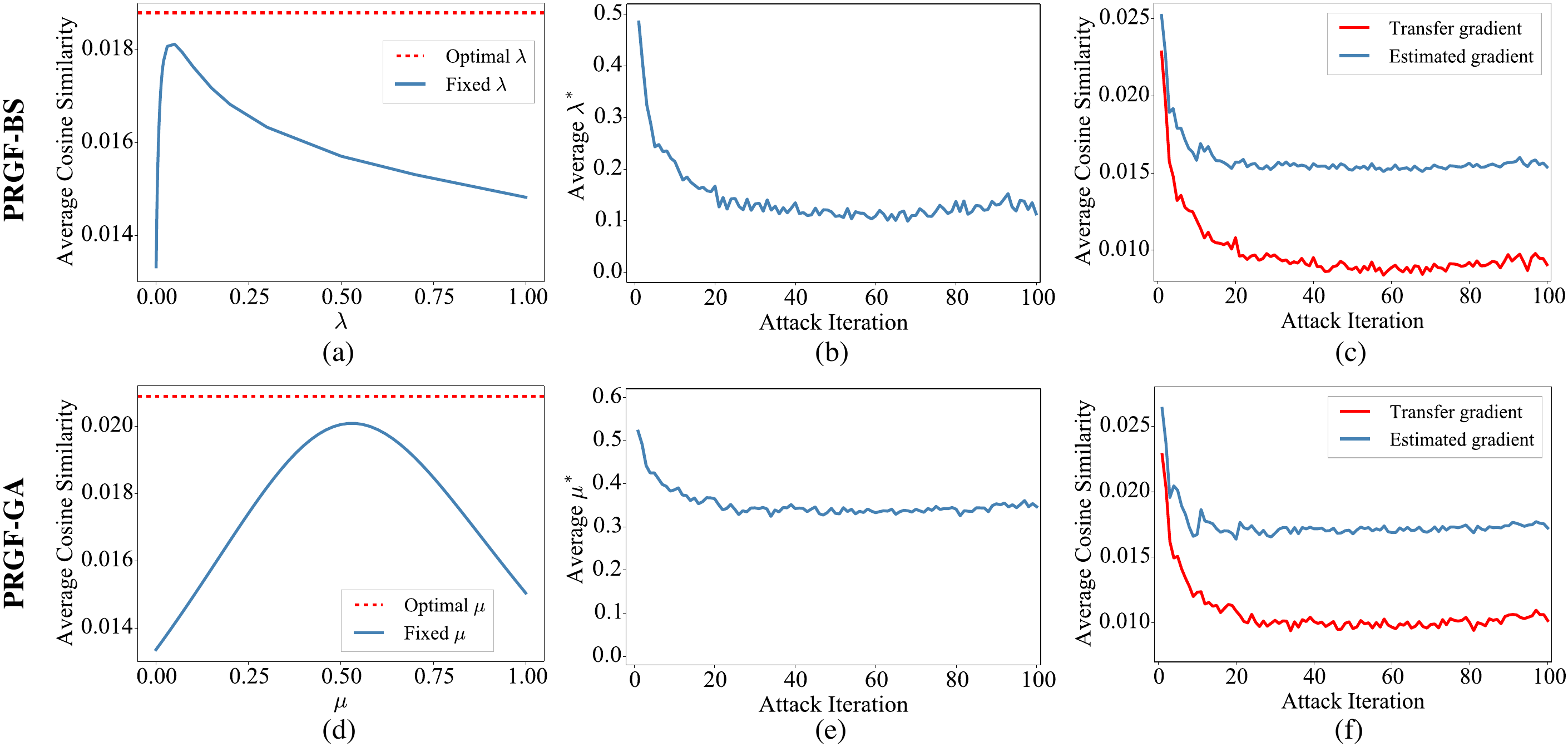}
\vspace{-.3cm}
\caption{(a) The average cosine similarity between the estimated gradient and the true gradient. The estimate is given by PRGF-BS with fixed $\lambda$ and optimal $\lambda$, respectively. (b) The average $\lambda^*$ in PRGF-BS across attack iterations. (c) The average cosine similarity between the transfer and the true gradients, and that between the estimated and the true gradients, across attack iterations in PRGF-BS. (d) The average cosine similarity between the estimated gradient and the true gradient. The estimate is given by PRGF-GA with fixed $\mu$ and optimal $\mu$, respectively. (e) The average $\mu^*$ in PRGF-GA across attack iterations. (f) The average cosine similarity between the transfer and the true gradients, and that between the estimated and the true gradients, across attack iterations in PRGF-GA.}
\label{fig:demo}
\vspace{-.2cm}
\end{figure*}
\begin{figure}[t]
\centering
\includegraphics[width=0.65\columnwidth]{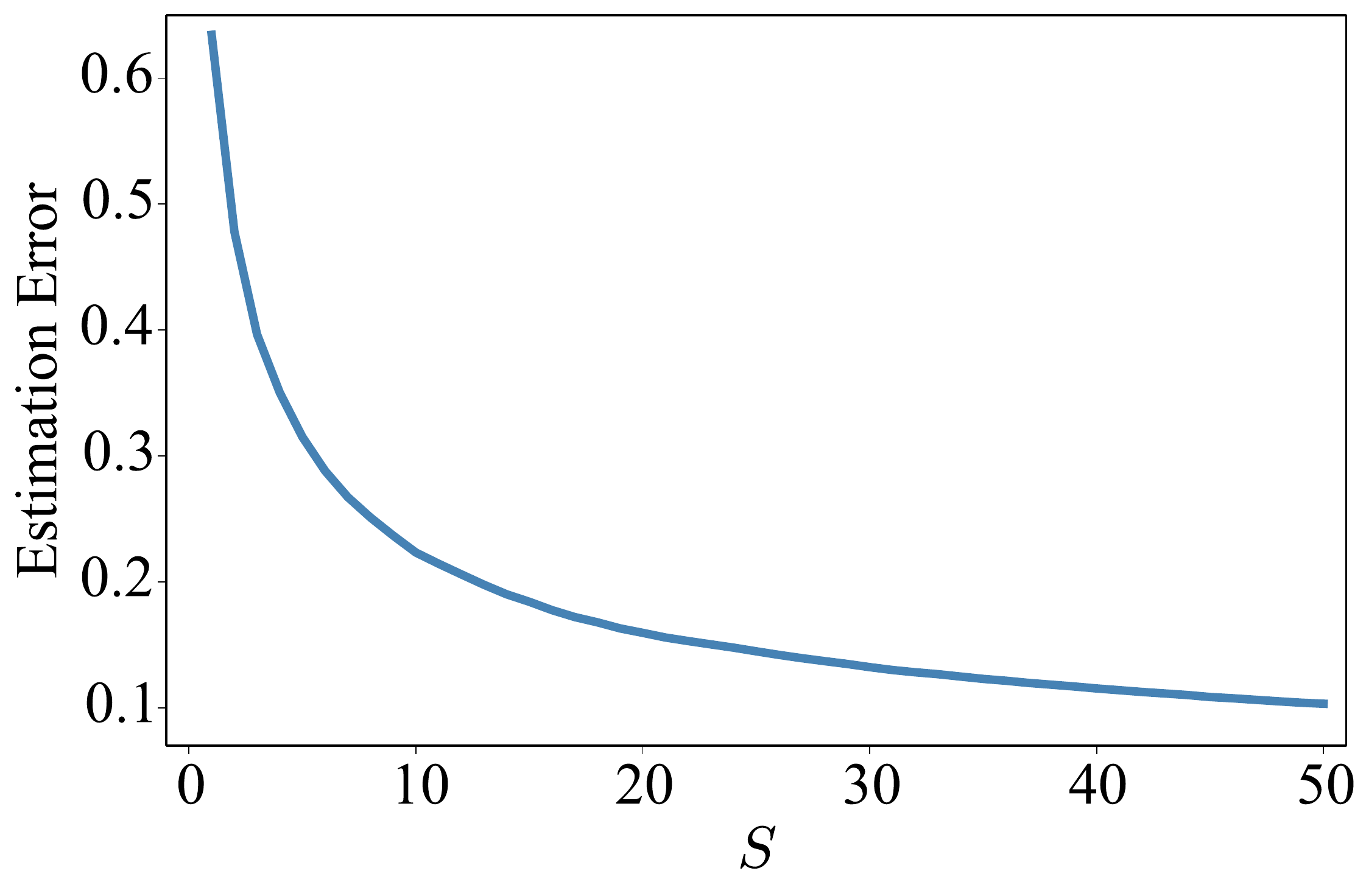}
\vspace{-.3cm}
\caption{The estimation error of gradient norm w.r.t. different queries $S$.}
\label{fig:errnorm}
\vspace{-.3cm}
\end{figure}

\subsection{Experimental Settings}\label{sec:6-0}

\textbf{ImageNet~\cite{russakovsky2015imagenet}.} We choose $1,000$ images randomly from the ILSVRC 2012 validation set to perform evaluations. Those images are normalized to $[0,1]$. 
We consider three black-box target models, which are Inception-v3~\cite{szegedy2016rethinking}, VGG-16~\cite{simonyan2014very}, and ResNet-50~\cite{He2015}.
For most experiments, we use the ResNet-v2-152 model~\cite{he2016identity} as the surrogate model to provide the transfer gradient. We also study different surrogate models in Section~\ref{sec:6-3-1}. 
For the proposed PRGF-BS and PRGF-GA algorithms, we set the number of queries in each step of gradient estimation as $q=50$ and the sampling variance as $\sigma=0.0001\cdot\sqrt{D}$.
We let the attack loss function $f$ in Eq.~\eqref{eq:problem} be the cross-entropy loss.
After we obtain the gradient estimate, we apply the PGD update rule as in Eq.~\eqref{eq:iter} to generate the adversarial example with the estimated gradient. We set the perturbation size as $\epsilon=\sqrt{0.001\cdot D}$ and the step size as $\eta=2$ in PGD under the $\ell_2$ norm, while set $\epsilon=0.05$ and $\eta=0.005$ under the $\ell_\infty$ norm.
For PRGF-GA, there is a preset threshold $c$ determining whether to directly return the transfer gradient, which is set as $c=\nicefrac{\sqrt{2}}{(\sqrt{2}+1)}$.

\textbf{CIFAR-10~\cite{krizhevsky2009learning}.} We adopt all the $10,000$ test images for evaluations, which are in $[0,1]$. The black-box target models include ResNet-50~\cite{He2015}, DenseNet-121~\cite{huang2017densely}, and SENet-18~\cite{hu2018squeeze}.
We adopt a Wide ResNet model (WRN-34-10)~\cite{zagoruyko2016wide} as the surrogate model.
We set $q=50$ and $\sigma=0.001\cdot\sqrt{D}$.
The loss function $f$ is the CW loss~\cite{carlini2016} since it performs better than the cross-entropy loss on CIFAR-10.
The perturbation size is $\epsilon=1.0$ under the $\ell_2$ norm and $\epsilon=\nicefrac{8}{255}$ under the $\ell_\infty$ norm. The step size in PGD is $\eta=0.25$ under the $\ell_2$ norm and $\eta=\nicefrac{2}{255}$ under the $\ell_\infty$ norm. The threshold in PRGF-GA is also set as $c=\nicefrac{\sqrt{2}}{(\sqrt{2}+1)}$.

Note that when counting the total number of queries in our methods, we include the additional queries of estimating the cosine similarity $\alpha$.

\subsection{Performance of Gradient Estimation}\label{sec:6-1}

We now conduct several ablation studies to show the performance of gradient estimation. All experiments in this section are performed on the Inception-v3~\cite{szegedy2016rethinking} model on ImageNet.

\textbf{Estimation of gradient norm.} First, we demonstrate the performance of gradient norm estimation as introduced in Section~\ref{sec:alpha}. In general, the gradient norm (or cosine similarity) is easier to estimate than the true gradient since it's a scalar value. Fig.~\ref{fig:errnorm} illustrates the estimation error of the gradient norm, defined as the (normalized) RMSE --- $\sqrt{\mathbb{E}\left(\frac{\widehat{\|\nabla f(x)\|_2} - \|\nabla f(x)\|_2 }{\|\nabla f(x)\|_2}\right)^2}$, w.r.t. the number of queries $S$, where $\|\nabla f(x)\|_2$ is the true norm, $\widehat{\|\nabla f(x)\|_2}$ is the estimated one, and the expectation is taken over all images along the attack procedure. 
It can be obtained that dozens of queries are sufficient to reach a small estimation error of gradient norm. We choose $S=10$ in the following experiments to reduce the number of queries while the estimation error is acceptable. 
The gradient norm is estimated every $10$ attack iterations to further reduce the required queries, since usually its value is relatively stable in the optimization process.

\begin{table*}
  \caption{The experimental results of black-box attacks against Inception-v3, VGG-16, and ResNet-50 under the $\ell_2$ norm on ImageNet. We report the attack success rate (ASR), and the average/median number of queries (AVG. Q/MED. Q) needed to generate an adversarial example over successful attacks. We mark the best results in \textbf{bold}. The subscript ``D'' denotes the methods with the data-dependent prior.}\vspace{-.2cm}
  \label{tab:final-results-full}
  \centering
  \begin{tabular}{l|ccc|ccc|ccc}
    \hline
    \multirow{2}{*}{Methods} & \multicolumn{3}{c|}{Inception-v3} & \multicolumn{3}{c|}{VGG-16} & \multicolumn{3}{c}{ResNet-50}\\
    \cline{2-10}
    & ASR & AVG. Q & MED. Q & ASR & AVG. Q & MED. Q & ASR & AVG. Q & MED. Q \\
    \hline
    NES~\cite{ilyas2018black} & 95.5\% & 1752 & 1071 & 98.7\% & 1103 & 816 & 98.4\% & 988 & 714 \\
    SPSA~\cite{uesato2018adversarial} & 93.7\% & 1808 & 1122 & 98.1\% & 1290 & 1020 & 98.4\% & 1236 & 969 \\
    AutoZoom~\cite{tu2018autozoom} & 85.4\% & 2443 & 1847 & 96.3\% & 1589 & 949 & 94.8\% & 2065 & 1223 \\
    Bandits\textsubscript{T}~\cite{ilyas2018prior} & 92.4\% & 1560 & 810 & 94.0\% & 584 & 225 & 96.2\% & 1076 & 446 \\
    Bandits\textsubscript{TD}~\cite{ilyas2018prior} & 97.2\% & 874 & 352 & 94.9\% & 278 & \bf82 & 96.8\% & 512 & 195 \\
    $\mathcal{N}$ATTACK~\cite{li2019nattack} & 98.2\% & 1020 & 510 & 99.6\% & 593 & 357 & 99.5\% & 535 & 357 \\
    \hline
    RGF & 97.7\% & 1309 & 816 & 99.8\% & 749 & 561 & 99.6\% & 673 & 510 \\
    PRGF-BS ($\lambda=0.05$) & 97.4\% & 1047 & 561 & 99.7\% & 624 & 408 & 99.3\% & 511 & 306 \\
    PRGF-BS ($\lambda^*$) & 98.1\% & 745 & 320 & 99.6\% & 331 & 182 & 99.6\% & 265 & \bf132 \\
    PRGF-GA ($\mu=0.5$) & 97.9\% & 958 & 572 & 99.8\% & 528 & 364 & 99.6\% & 485 & 312 \\
    PRGF-GA ($\mu^*$) & 97.9\% & 735 & 314 & 99.7\% & 320 & 184 & 99.5\% & 250 & 134 \\
    \hline
    RGF\textsubscript{D} & 99.1\% & 910 & 561 & \bf100.0\% & 372 & 306 & 99.7\% & 429 & 306 \\
    PRGF-BS\textsubscript{D} ($\lambda=0.05$) & 98.8\% & 728 & 408 & 99.9\% & 359 & 255 & \bf99.8\% & 379 & 255 \\
    PRGF-BS\textsubscript{D} ($\lambda^*$) & 99.1\% & 649 & 332 & 99.8\% & 250 & 180 & 99.6\% & \bf232 & 140 \\
    PRGF-GA\textsubscript{D} ($\mu=0.5$) & \bf99.3\% & 734 & 416 & \bf100.0\% & 332 & 260 & 99.7\% & 340 & 260 \\
    PRGF-GA\textsubscript{D} ($\mu^*$) & 99.2\% & \bf644 & \bf312 & 99.7\% & \bf239 & 184 & 99.7\% & 240 & 140 \\
    \hline
  \end{tabular}
\end{table*}

\textbf{Performance of gradient estimation.} Second, we verify the effectiveness of the derived \emph{optimal} $\lambda$ in PRGF-BS and $\mu$ in PRGF-GA (i.e., $\lambda^*$ in Eq.~\eqref{eq:lambda-1} and $\mu^*$ in Eq.~\eqref{eq:mu-1}) for gradient estimation, compared with any fixed $\lambda, \mu\in[0,1]$. 
To this end, we perform attacks against Inception-v3 using PRGF-BS with $\lambda^*$ or PRGF-GA with $\mu^*$, and at the same time calculate the cosine similarity between the estimated gradient and the true gradient.
In both methods, $\lambda^*$ and $\mu^*$ are calculated using the estimated $\alpha$ instead of its true value.
Meanwhile, along the PGD updates, we also use fixed $\lambda$ or $\mu$ to get gradient estimates, and calculate the corresponding cosine similarities. Note that $\lambda^*$ and $\mu^*$ do not correspond to any fixed value, since they vary during iterations. 

We show the average cosine similarities of different fixed values of $\lambda$ in Fig.~\ref{fig:demo}(a), and those of different fixed values of $\mu$ in Fig.~\ref{fig:demo}(d).
The first observation is that when a suitable value of $\lambda$ (or $\mu$) is chosen, the proposed PRGF-BS (or PRGF-GA) provides a better gradient estimate than both the ordinary RGF method with uniform distribution (when $\lambda=\frac{1}{D}\approx 0$ or $\mu=0$) and the transfer gradient (when $\lambda=1$ or $\mu=1$). The second observation is that adopting $\lambda^*$ (or $\mu^*$) brings further improvement upon any fixed $\lambda$ (or $\mu$), demonstrating the effectiveness of our theoretical analyses.

\textbf{Gradient estimation across attack iterations.} Finally, we aim at examining the effectiveness of the transfer-based prior across attack iterations. We show the average $\lambda^*$ and $\mu^*$ over all images w.r.t. attack iterations in Fig.~\ref{fig:demo}(b) for PRGF-BS, and in Fig.~\ref{fig:demo}(e) for PRGF-GA, respectively. The curves show that $\lambda^*$ and $\mu^*$ decrease along the iterations. Besides, Fig.~\ref{fig:demo}(c) and Fig.~\ref{fig:demo}(f) show the average cosine similarity between the transfer and the true gradients, and that between the estimated and the true gradients w.r.t. attack iterations, in PRGF-BS and PRGF-GA. All of these results demonstrate that the transfer gradient is more useful at beginning, and becomes less useful along the iterations. However, the estimated gradient in either PRGF-BS or PRGF-GA can remain a higher cosine similarity with the true gradient, which facilitates the adversarial attacks consequently. The results also corroborate that we need to use the adaptive $\lambda^*$ or $\mu^*$ in different attack iterations.

\subsection{Results on ImageNet}\label{sec:6-2}

\begin{table*}
  \caption{The experimental results of PRGF-BS and PRGF-GA attacks against Inception-v3, VGG-16, and ResNet-50 under the $\ell_2$ norm on ImageNet using different surrogate models. We report the attack success rate (ASR), and the average/median number of queries (AVG. Q/MED. Q) needed to generate an adversarial example over successful attacks. We mark the best results in \textbf{bold}. The subscript ``D'' denotes the methods with the data-dependent prior.}
  \label{tab:surrogate-models}\vspace{-.2cm}
  \centering
  \begin{tabular}{c|l|ccc|ccc|ccc}
    \hline
    \multirow{2}{*}{Surrogate Model(s)} & \multirow{2}{*}{Methods} & \multicolumn{3}{c|}{Inception-v3} & \multicolumn{3}{c|}{VGG-16} & \multicolumn{3}{c}{ResNet-50}\\
    \cline{3-11}
   & & ASR & AVG. Q & MED. Q & ASR & AVG. Q & MED. Q & ASR & AVG. Q & MED. Q \\
    \hline
    \multirow{4}{*}{ResNet-v2-152} & PRGF-BS & 98.1\% & 745 & 320 & 99.6\% & 331 & 182 & 99.6\% & 265 & 132 \\
    & PRGF-GA & 97.9\% & 735 & 314 & 99.7\% & 320 & 184 & 99.5\% & 250 & 134 \\
    & PRGF-BS\textsubscript{D} & 99.1\% & 649 & 332 & 99.8\% & 250 & 180 & 99.6\% & 232 & 140 \\
    & PRGF-GA\textsubscript{D} & 99.2\% & 644 & 312 & 99.7\% & 239 & 184 & 99.7\% & 240 & 140 \\
    \hline
    \multirow{4}{*}{Inception-v4} & PRGF-BS & 98.9\% & 673 & 252 & 99.9\% & 350 & 184 & 99.7\% & 386 & 234 \\
    & PRGF-GA & 99.0\% & 622 & 242 & 99.9\% & 343 & 186 & 99.7\% & 350 & 192 \\
    & PRGF-BS\textsubscript{D} & 99.2\% & 569 & 252 & 99.9\% & 251 & 180 & 99.7\% & 298 & 224 \\
    & PRGF-GA\textsubscript{D} & 99.5\% & 595 & 248 & \bf100.0\% & 256 & 186 & 99.8\% & 298 & 194 \\
    \hline
    \multirow{4}{*}{\makecell{ ResNet-v2-152 \\ + Inception-v4 \\ (equal averaging) } } & PRGF-BS & 98.2\% & 592 & 188 & 99.7\% & 290 & 132 & 99.8\% & 282 & 130 \\
    & PRGF-GA & 98.7\% & 575 & 190 & 99.9\% & 283 & 134 & 99.7\% & 262 & 132 \\
    & PRGF-BS\textsubscript{D} & 99.2\% & 537 & 230 & 99.9\% & 219 & 140 & 99.7\% & 245 & 138 \\
    & PRGF-GA\textsubscript{D} & 99.1\% & 516 & 236 & 99.9\% & 219 & 136 & 99.7\% & 242 & 138 \\
    \hline
    \multirow{4}{*}{\makecell{ ResNet-v2-152 \\ + Inception-v4 \\ (subspace projection) } } & PRGF-BS & 99.1\% & 348 & 94 & 99.9\% & 163 & 59 & 99.6\% & 151 & 70 \\
    & PRGF-GA & 99.5\% & 342 & 96 & \bf100.0\% & 146 & 66 & 99.6\% & 135 & 70 \\
    & PRGF-BS\textsubscript{D} & 99.1\% & 412 & 156 & 99.9\% & 165 & 94 & 99.8\% & 182 & 105 \\
    & PRGF-GA\textsubscript{D} & 99.5\% & 404 & 152 & \bf100.0\% & 169 & 96 & 99.7\% & 164 & 96 \\
    \hline
    \multirow{4}{*}{\makecell{ ResNet-v2-152 \\ + Inception-v4 \\ + Inception-ResNet-v2 \\ (subspace projection) } } & PRGF-BS & 99.5\% & 198 & \bf50 & \bf100.0\% & 93 & \bf40 & \bf99.9\% & 103 & \bf40 \\
    & PRGF-GA & \bf99.8\% & \bf191 & \bf50 & \bf100.0\% & \bf89 & \bf40 & 99.8\% & \bf96 & \bf40 \\
    & PRGF-BS\textsubscript{D} & 99.7\% & 296 & 95 & 99.9\% & 122 & 70 & \bf99.9\% & 135 & 75 \\
    & PRGF-GA\textsubscript{D} & 99.6\% & 267 & 97 & \bf100.0\% & 118 & 72 & 99.8\% & 126 & 77 \\
    \hline
  \end{tabular}
  \vspace{-1.ex}
\end{table*}

In this section, we perform black-box adversarial attacks against three ImageNet models, including Inception-v3~\cite{szegedy2016rethinking}, VGG-16~\cite{simonyan2014very}, and ResNet-50~\cite{He2015}. 
Besides the two proposed PRGF-BS and PRGF-GA algorithms, 
we incorporate several baseline methods, including the ordinary RGF method with uniform sampling, the PRGF-BS method with the fixed $\lambda=0.05$, and the PRGF-GA method with the fixed $\mu=0.5$. Those fixed values are chosen according to Fig.~\ref{fig:demo}(a) and Fig.~\ref{fig:demo}(d), which can estimate the gradient more accurately.
We set the number of queries as $q=50$ for gradient estimation and the sampling variance as $\sigma=0.0001\cdot\sqrt{D}$, which are identical for all of these methods.
We also incorporate the data-dependent prior into these methods for comparison (which are denoted by adding a subscript ``D''). We set the dimension of the subspace as $d=50\times50\times3$. 

Besides, we compare the attack performance with various state-of-the-art attack methods, including the natural evolution strategies (NES)~\cite{ilyas2018black}, SPSA~\cite{uesato2018adversarial}, AutoZoom~\cite{tu2018autozoom}, bandit optimization methods (Bandits\textsubscript{T} and Bandits\textsubscript{TD})~\cite{ilyas2018prior}, and $\mathcal{N}$ATTACK~\cite{li2019nattack}.
For all methods, we restrict the maximum number of queries for each image to be $10$,$000$.
We report a successful attack if a method can generate an adversarial example within $10$,$000$ queries and the size of perturbation is smaller than the budget (i.e., $\epsilon=\sqrt{0.001\cdot D}$).

\begin{table*}
  \caption{The experimental results of black-box attacks against ResNet-50, DenseNet-121, and SENet-18 under the $\ell_2$ norm on CIFAR-10. We report the attack success rate (ASR) and the average/median number of queries (AVG. Q/MED. Q) needed to generate an adversarial example over successful attacks. We mark the best results in \textbf{bold}.}
  \label{tab:cifar-l2}
\vspace{-.2cm}
  \centering
  \begin{tabular}{l|ccc|ccc|ccc}
    \hline
    \multirow{2}{*}{Methods} & \multicolumn{3}{c|}{ResNet-50} & \multicolumn{3}{c|}{DenseNet-121} & \multicolumn{3}{c}{SENet-18}\\
    \cline{2-10}
    & ASR & AVG. Q & MED. Q & ASR & AVG. Q & MED. Q & ASR & AVG. Q & MED. Q \\
    \hline
    NES~\cite{ilyas2018black} & 99.7\% & 642 & 459 & 99.6\% & 631 & 459 & 99.8\% & 582 & 408 \\
    SPSA~\cite{uesato2018adversarial} & 99.8\% & 785 & 561 & 99.7\% & 780 & 510 & 99.9\% & 718 & 459 \\
    Bandits\textsubscript{T}~\cite{ilyas2018prior} & \bf100.0\% & 375 & 194 & \bf100.0\% & 356 & 174 & \bf100.0\% & 317 & 150 \\
    $\mathcal{N}$ATTACK~\cite{li2019nattack} & \bf100.0\% & 401 & 255 & \bf100.0\% & 404 & 255 & \bf100.0\% & 350 & 204 \\
    \hline
    RGF & 99.9\% & 460 & 357 & 99.9\% & 472 & 357 & 99.9\% & 423 & 306 \\
    PRGF-BS ($\lambda=0.05$) & 99.8\% & 290 & 204 & 99.9\% & 274 & 204 & 99.9\% & 262 & 153 \\
    PRGF-BS ($\lambda^*$) & 99.9\% & 268 & 124 & \bf100.0\% & 220 & 124 & \bf100.0\% & 187 & 76 \\
    PRGF-GA ($\mu=0.5$) & 99.3\% & 306 & 204 & 99.7\% & 260 & 204 & 99.9\% & 243 & 153 \\
    PRGF-GA ($\mu^*$) & 99.9\% & \bf173 & \bf76 & 99.9\% & \bf168 & \bf76 & 99.9\% & \bf146 & \bf65 \\
    \hline
  \end{tabular}
  \vspace{-1ex}
\end{table*}

\begin{figure*}[t]
\centering
\includegraphics[width=0.95\linewidth]{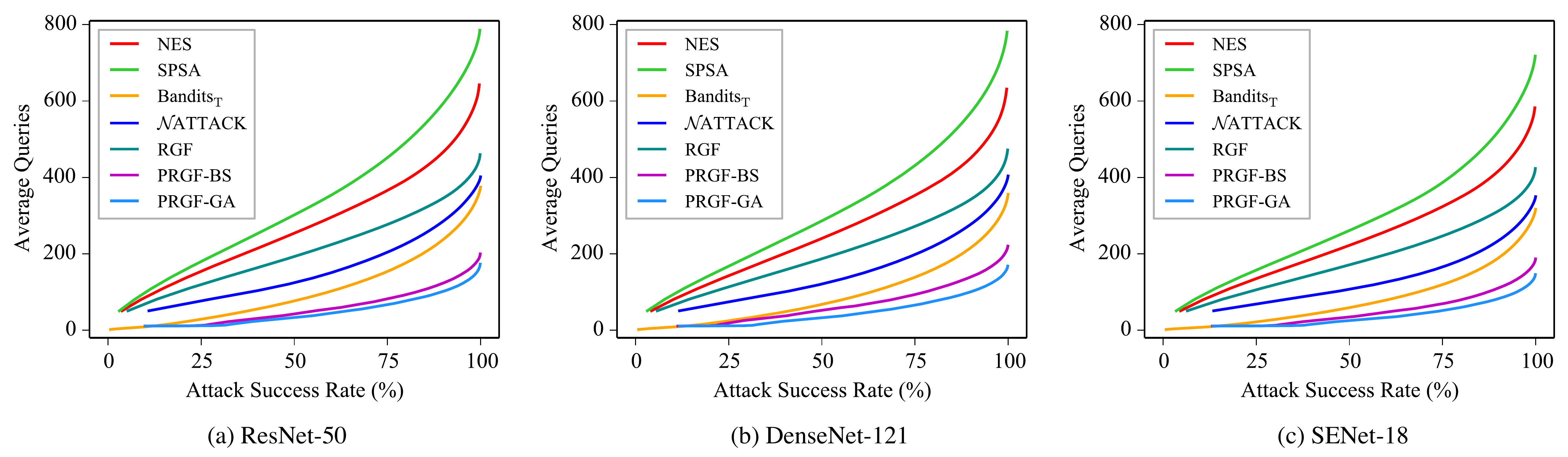}
\vspace{-.2cm}
\caption{The average number of queries for generating the adversarial examples that are successfully misclassified by the black-box model at any desired success rate on CIFAR-10.}
\label{fig:curve}\vspace{-.2cm}
\end{figure*}

Table~\ref{tab:final-results-full} shows the results, where we report the success rate of black-box attacks and the average/median number of queries needed to generate an adversarial example over successful attacks.
We have the following observations.
First, compared with the state-of-the-art attacks, the proposed methods generally lead to higher attack success rates and require much fewer queries.
Second, the transfer-based prior provides useful prior information for black-box attacks since PRGF based methods perform better than the ordinary RGF method.
Third, using a fixed $\lambda$ in PRGF-BS or a fixed $\mu$ in PRGF-GA cannot exceed the performance of using their optimal values, although they already lead to comparable performance with the state-of-the-art methods. 
Fourth, the results also prove that the data-dependent prior is orthogonal to the proposed transfer-based prior, since integrating the data-dependent prior leads to better results.
Fifth, PRGF-GA requires slightly fewer queries than PRGF-BS in most cases, which are consistent with the loss curves in Fig.~\ref{fig:theory}.

\subsubsection{Different Surrogate Models}\label{sec:6-3-1}

Here we conduct an ablation study to investigate the effectiveness of adopting different surrogate models. We use the ResNet-v2-152 model~\cite{he2016identity} as the the surrogate model in the above experiments. We additionally consider Inception-v4~\cite{szegedy2017inception}, ResNet-v2-152 + Inception-v4, and ResNet-v2-152 + Inception-v4 + Inception-ResNet-v2~\cite{szegedy2017inception} as the surrogate models. Note that the latter two include multiple surrogate models. We adopt the \emph{subspace projection} method introduced in Section~\ref{sec:5-2} to get the transfer-based prior when multiple surrogate models are available. Besides, we also compare this method with the \emph{equal averaging} method that directly averages the gradients of multiple models (only in the case of using ResNet-v2-152 + Inception-v4).

We show the attack performance of PRGF-BS, PRGF-GA, PRGF-BS\textsubscript{D}, and PRGF-GA\textsubscript{D} with different surrogate models in Table~\ref{tab:surrogate-models}.
It is easy to see that adopting multiple surrogate models can significantly improve the attack success rates and reduce the number of queries. When using three surrogate models, the median number of queries is less than $100$ for all target models, which validates the effectiveness of the transfer-based prior. Besides, it can be noted that the subspace projection method performs better than the equal averaging method, because the subspace projection method can obtain the transfer-based prior which approximates the true gradient best in the subspace.

Another observation from the results is that adopting a similar surrogate model of the target model can enhance the attack performance. In particular, ResNet-v2-152 is better than Inception-v4 as the surrogate model for attacking the ResNet-50 models. On the other hand, Inception-v4 is better than ResNet-v2-152 for attacking the Inception-v3 model.
It is reasonable since the gradients of models within the same family of model architectures would be similar, which has been verified in~\cite{su2018robustness} showing that the adversarial transferability is higher across similar model architectures.

Finally, we find that the data-dependent prior becomes less useful with a more powerful transfer-based prior obtained by multiple surrogate models. Specifically, PRGF-BS\textsubscript{D} and PRGF-GA\textsubscript{D} require more queries than PRGF-BS and PRGF-GA when using two or three surrogate models. The reason is as follows. For PRGF-BS and PRGF-GA without the data-dependent prior, it is more likely to obtain $\lambda^*=1$ or $\mu^*=1$ with the more effective transfer-based prior, such that we do not need to perform $q$ queries to estimate the gradient. However, in PRGF-BS\textsubscript{D} and PRGF-GA\textsubscript{D}, $\lambda^*$ and $\mu^*$ are less probable to be $1$ due to that sampling in the data-dependent subspace can also improve the gradient estimate, and therefore we need $q$ more queries to get the estimate. 
Although the data-dependent prior helps to give a more accurate gradient estimate, the cost of $q$ more queries degrades the efficiency of attacks.

\subsection{Results on CIFAR-10}\label{sec:6-3}

In this section, we show the results of black-box adversarial attacks on CIFAR-10. Similar to the experiments on ImageNet, we compare the performance of PRGF-BS and PRGF-GA with three baselines --- RGF, PRGF-BS with the fixed $\lambda=0.05$, and PRGF-GA with the fixed $\mu=0.5$, as well as four other attacks --- NES~\cite{ilyas2018black}, SPSA~\cite{uesato2018adversarial}, Bandits\textsubscript{T}~\cite{ilyas2018prior}, and $\mathcal{N}$ATTACK~\cite{li2019nattack}.
Since the image resolution in CIFAR-10 is not very high (i.e., $32\times32\times3$), we do not adopt the data-dependent prior.
We also restrict the maximum number of queries for each image to be $10,000$. Note that hundreds of queries could be sufficient due to the lower input dimension of CIFAR-10, but we adopt the maximum $10,000$ queries to make it consistent with the setting on ImageNet.

The black-box attack results of those methods against ResNet-50~\cite{He2015}, DenseNet-121~\cite{huang2017densely}, and SENet-18~\cite{hu2018squeeze} are presented in Table~\ref{tab:cifar-l2}. It can be seen that with the maximum number of $10,000$ queries, all attack methods can achieve near $100\%$ attack success rate. Nevertheless, our proposed methods (especially PRGF-GA) require much less queries to successfully generate adversarial examples, which demonstrates the query efficiency of our proposed methods.

Fig.~\ref{fig:curve} shows the average number of queries for successfully misleading the black-box model by reaching a desired success rate. For a given attack success rate, our methods require much less queries, indicating that they are much more query-efficient than other baseline methods.

\section{Conclusion}\label{sec:7}
In this paper, two prior-guided random gradient-free algorithms were proposed for improving black-box attacks. Our methods can utilize a transfer-based prior given by the gradient of a surrogate model through biased sampling and gradient averaging, respectively.
We appropriately integrated the transfer-based prior with model queries by the derived optimal coefficient in both methods under the gradient estimation framework.
Furthermore, we extended the proposed methods by incorporating the data-dependent prior and utilizing multiple surrogate models.
The experimental results consistently demonstrate the effectiveness of our methods, which require much fewer queries to attack black-box models with higher success rates compared with various state-of-the-art attack methods. 
We released our codes at \url{https://github.com/thu-ml/Prior-Guided-RGF}.


%



\ifCLASSOPTIONcompsoc
  \section*{Acknowledgments}
\else
  \section*{Acknowledgment}
\fi

\small
This work was supported by the National Key Research and Development Program of China (No. 2020AAA0104304), NSFC Projects (Nos. 61620106010, 62061136001, 61621136008, 62076147, U19B2034, U1811461, U19A2081), Beijing NSF Project (No. JQ19016), Beijing Academy of Artificial Intelligence (BAAI),  Tsinghua-Huawei Joint Research Program, Tsinghua Institute for Guo Qiang, Tsinghua-OPPO Joint Research Center for Future Terminal Technology and Tsinghua-China Mobile Communications Group Co., Ltd. Joint Institute.


\ifCLASSOPTIONcaptionsoff
  \newpage
\fi



%
\bibliographystyle{ieeetr}
\bibliography{egbib}

\begin{thebibliography}{10}

\bibitem{Goodfellow-et-al2016}
I.~Goodfellow, Y.~Bengio, and A.~Courville, {\em Deep Learning}.
\newblock MIT Press, 2016.
\newblock \url{http://www.deeplearningbook.org}.

\bibitem{Biggio2013Evasion}
B.~Biggio, I.~Corona, D.~Maiorca, B.~Nelson, P.~Laskov, G.~Giacinto, and
  F.~Roli, ``Evasion attacks against machine learning at test time,'' in {\em
  Joint European Conference on Machine Learning and Knowledge Discovery in
  Databases}, pp.~387--402, 2013.

\bibitem{Szegedy2013}
C.~Szegedy, W.~Zaremba, I.~Sutskever, J.~Bruna, D.~Erhan, I.~Goodfellow, and
  R.~Fergus, ``Intriguing properties of neural networks,'' in {\em
  International Conference on Learning Representations (ICLR)}, 2014.

\bibitem{Goodfellow2014}
I.~J. Goodfellow, J.~Shlens, and C.~Szegedy, ``Explaining and harnessing
  adversarial examples,'' in {\em International Conference on Learning
  Representations (ICLR)}, 2015.

\bibitem{Kurakin2016}
A.~Kurakin, I.~Goodfellow, and S.~Bengio, ``Adversarial examples in the
  physical world,'' {\em arXiv preprint arXiv:1607.02533}, 2016.

\bibitem{Athalye2018Obfuscated}
A.~Athalye, N.~Carlini, and D.~Wagner, ``Obfuscated gradients give a false
  sense of security: Circumventing defenses to adversarial examples,'' in {\em
  International Conference on Machine Learning (ICML)}, pp.~274--283, 2018.

\bibitem{dong2019evading}
Y.~Dong, T.~Pang, H.~Su, and J.~Zhu, ``Evading defenses to transferable
  adversarial examples by translation-invariant attacks,'' in {\em Proceedings
  of the IEEE Conference on Computer Vision and Pattern Recognition (CVPR)},
  pp.~4312--4321, 2019.

\bibitem{carlini2016}
N.~Carlini and D.~Wagner, ``Towards evaluating the robustness of neural
  networks,'' in {\em IEEE Symposium on Security and Privacy}, pp.~39--57,
  2017.

\bibitem{madry2017towards}
A.~Madry, A.~Makelov, L.~Schmidt, D.~Tsipras, and A.~Vladu, ``Towards deep
  learning models resistant to adversarial attacks,'' in {\em International
  Conference on Learning Representations (ICLR)}, 2018.

\bibitem{dong2020benchmarking}
Y.~Dong, Q.-A. Fu, X.~Yang, T.~Pang, H.~Su, Z.~Xiao, and J.~Zhu, ``Benchmarking
  adversarial robustness on image classification,'' in {\em Proceedings of the
  IEEE/CVF Conference on Computer Vision and Pattern Recognition (CVPR)},
  pp.~321--331, 2020.

\bibitem{pang2021tricks-at}
T.~Pang, X.~Yang, Y.~Dong, H.~Su, and J.~Zhu, ``Bag of tricks for adversarial
  training,'' in {\em International Conference on Learning Representations
  (ICLR)}, 2021.

\bibitem{ilyas2018black}
A.~Ilyas, L.~Engstrom, A.~Athalye, and J.~Lin, ``Black-box adversarial attacks
  with limited queries and information,'' in {\em International Conference on
  Machine Learning (ICML)}, pp.~2137--2146, 2018.

\bibitem{Brendel2018Decision}
W.~Brendel, J.~Rauber, and M.~Bethge, ``Decision-based adversarial attacks:
  Reliable attacks against black-box machine learning models,'' in {\em
  International Conference on Learning Representations (ICLR)}, 2018.

\bibitem{papernot2016practical}
N.~Papernot, P.~McDaniel, I.~Goodfellow, S.~Jha, Z.~B. Celik, and A.~Swami,
  ``Practical black-box attacks against machine learning,'' in {\em Proceedings
  of the 2017 ACM on Asia Conference on Computer and Communications Security},
  pp.~506--519, 2017.

\bibitem{Chen2017ZOO}
P.~Y. Chen, H.~Zhang, Y.~Sharma, J.~Yi, and C.~J. Hsieh, ``Zoo: Zeroth order
  optimization based black-box attacks to deep neural networks without training
  substitute models,'' in {\em ACM Workshop on Artificial Intelligence and
  Security (AISec)}, pp.~15--26, 2017.

\bibitem{Dong2017}
Y.~Dong, F.~Liao, T.~Pang, H.~Su, J.~Zhu, X.~Hu, and J.~Li, ``Boosting
  adversarial attacks with momentum,'' in {\em Proceedings of the IEEE
  Conference on Computer Vision and Pattern Recognition (CVPR)},
  pp.~9185--9193, 2018.

\bibitem{Bhagoji_2018_ECCV}
A.~Nitin~Bhagoji, W.~He, B.~Li, and D.~Song, ``Practical black-box attacks on
  deep neural networks using efficient query mechanisms,'' in {\em Proceedings
  of the European Conference on Computer Vision (ECCV)}, pp.~154--169, 2018.

\bibitem{tu2018autozoom}
C.-C. Tu, P.~Ting, P.-Y. Chen, S.~Liu, H.~Zhang, J.~Yi, C.-J. Hsieh, and S.-M.
  Cheng, ``Autozoom: Autoencoder-based zeroth order optimization method for
  attacking black-box neural networks,'' in {\em Proceedings of the
  Thirty-Third AAAI Conference on Artificial Intelligence (AAAI)},
  pp.~742--749, 2019.

\bibitem{ilyas2018prior}
A.~Ilyas, L.~Engstrom, and A.~Madry, ``Prior convictions: Black-box adversarial
  attacks with bandits and priors,'' in {\em International Conference on
  Learning Representations (ICLR)}, 2019.

\bibitem{dong2019efficient}
Y.~Dong, H.~Su, B.~Wu, Z.~Li, W.~Liu, T.~Zhang, and J.~Zhu, ``Efficient
  decision-based black-box adversarial attacks on face recognition,'' in {\em
  Proceedings of the IEEE Conference on Computer Vision and Pattern Recognition
  (CVPR)}, pp.~7714--7722, 2019.

\bibitem{Papernot20162}
N.~Papernot, P.~McDaniel, and I.~Goodfellow, ``Transferability in machine
  learning: from phenomena to black-box attacks using adversarial samples,''
  {\em arXiv preprint arXiv:1605.07277}, 2016.

\bibitem{Liu2016}
Y.~Liu, X.~Chen, C.~Liu, and D.~Song, ``Delving into transferable adversarial
  examples and black-box attacks,'' in {\em International Conference on
  Learning Representations (ICLR)}, 2017.

\bibitem{xie2019improving}
C.~Xie, Z.~Zhang, Y.~Zhou, S.~Bai, J.~Wang, Z.~Ren, and A.~L. Yuille,
  ``Improving transferability of adversarial examples with input diversity,''
  in {\em Proceedings of the IEEE Conference on Computer Vision and Pattern
  Recognition (CVPR)}, pp.~2730--2739, 2019.

\bibitem{nesterov2017random}
Y.~Nesterov and V.~Spokoiny, ``Random gradient-free minimization of convex
  functions,'' {\em Foundations of Computational Mathematics}, vol.~17, no.~2,
  pp.~527--566, 2017.

\bibitem{ghadimi2013stochastic}
S.~Ghadimi and G.~Lan, ``Stochastic first-and zeroth-order methods for
  nonconvex stochastic programming,'' {\em SIAM Journal on Optimization},
  vol.~23, no.~4, pp.~2341--2368, 2013.

\bibitem{duchi2015optimal}
J.~C. Duchi, M.~I. Jordan, M.~J. Wainwright, and A.~Wibisono, ``Optimal rates
  for zero-order convex optimization: The power of two function evaluations,''
  {\em IEEE Transactions on Information Theory}, vol.~61, no.~5,
  pp.~2788--2806, 2015.

\bibitem{cheng2019improving}
S.~Cheng, Y.~Dong, T.~Pang, H.~Su, and J.~Zhu, ``Improving black-box
  adversarial attacks with a transfer-based prior,'' in {\em Advances in Neural
  Information Processing Systems (NeurIPS)}, pp.~10934--10944, 2019.

\bibitem{lax2014calculus}
P.~D. Lax and M.~S. Terrell, {\em Calculus with applications}.
\newblock Springer, 2014.

\bibitem{russakovsky2015imagenet}
O.~Russakovsky, J.~Deng, H.~Su, J.~Krause, S.~Satheesh, S.~Ma, Z.~Huang,
  A.~Karpathy, A.~Khosla, M.~Bernstein, {\em et~al.}, ``Imagenet large scale
  visual recognition challenge,'' {\em International Journal of Computer
  Vision}, vol.~115, no.~3, pp.~211--252, 2015.

\bibitem{JMLR:v15:wierstra14a}
D.~Wierstra, T.~Schaul, T.~Glasmachers, Y.~Sun, J.~Peters, and J.~Schmidhuber,
  ``Natural evolution strategies,'' {\em Journal of Machine Learning Research},
  vol.~15, no.~27, pp.~949--980, 2014.

\bibitem{oh2017towards}
S.~J. Oh, M.~Augustin, B.~Schiele, and M.~Fritz, ``Towards reverse-engineering
  black-box neural networks,'' in {\em International Conference on Learning
  Representations (ICLR)}, 2018.

\bibitem{brunner2019guessing}
T.~Brunner, F.~Diehl, M.~T. Le, and A.~Knoll, ``Guessing smart: Biased sampling
  for efficient black-box adversarial attacks,'' in {\em Proceedings of the
  IEEE International Conference on Computer Vision (ICCV)}, pp.~4958--4966,
  2019.

\bibitem{maheswaranathan2018guided}
N.~Maheswaranathan, L.~Metz, G.~Tucker, D.~Choi, and J.~Sohl-Dickstein,
  ``Guided evolutionary strategies: Augmenting random search with surrogate
  gradients,'' in {\em International Conference on Machine Learning (ICML)},
  pp.~4264--4273, 2019.

\bibitem{guo2018low}
C.~Guo, J.~S. Frank, and K.~Q. Weinberger, ``Low frequency adversarial
  perturbation,'' {\em arXiv preprint arXiv:1809.08758}, 2018.

\bibitem{su2018robustness}
D.~Su, H.~Zhang, H.~Chen, J.~Yi, P.-Y. Chen, and Y.~Gao, ``Is robustness the
  cost of accuracy?--a comprehensive study on the robustness of 18 deep image
  classification models,'' in {\em Proceedings of the European Conference on
  Computer Vision (ECCV)}, pp.~631--648, 2018.

\bibitem{krizhevsky2009learning}
A.~Krizhevsky and G.~Hinton, ``Learning multiple layers of features from tiny
  images,'' tech. rep., University of Toronto, 2009.

\bibitem{szegedy2016rethinking}
C.~Szegedy, V.~Vanhoucke, S.~Ioffe, J.~Shlens, and Z.~Wojna, ``Rethinking the
  inception architecture for computer vision,'' in {\em Proceedings of the IEEE
  Conference on Computer Vision and Pattern Recognition (CVPR)},
  pp.~2818--2826, 2016.

\bibitem{simonyan2014very}
K.~Simonyan and A.~Zisserman, ``Very deep convolutional networks for
  large-scale image recognition,'' in {\em International Conference on Learning
  Representations (ICLR)}, 2015.

\bibitem{He2015}
K.~He, X.~Zhang, S.~Ren, and J.~Sun, ``Deep residual learning for image
  recognition,'' in {\em Proceedings of the IEEE Conference on Computer Vision
  and Pattern Recognition (CVPR)}, pp.~770--778, 2016.

\bibitem{he2016identity}
K.~He, X.~Zhang, S.~Ren, and J.~Sun, ``Identity mappings in deep residual
  networks,'' in {\em Proceedings of the European Conference on Computer Vision
  (ECCV)}, pp.~630--645, 2016.

\bibitem{huang2017densely}
G.~Huang, Z.~Liu, L.~Van Der~Maaten, and K.~Q. Weinberger, ``Densely connected
  convolutional networks,'' in {\em Proceedings of the IEEE Conference on
  Computer Vision and Pattern Recognition (CVPR)}, pp.~4700--4708, 2017.

\bibitem{hu2018squeeze}
J.~Hu, L.~Shen, and G.~Sun, ``Squeeze-and-excitation networks,'' in {\em
  Proceedings of the IEEE Conference on Computer Vision and Pattern Recognition
  (CVPR)}, pp.~7132--7141, 2018.

\bibitem{zagoruyko2016wide}
S.~Zagoruyko and N.~Komodakis, ``Wide residual networks,'' in {\em Proceedings
  of the British Machine Vision Conference (BMVC)}, 2016.

\bibitem{uesato2018adversarial}
J.~Uesato, B.~O'Donoghue, A.~v.~d. Oord, and P.~Kohli, ``Adversarial risk and
  the dangers of evaluating against weak attacks,'' in {\em International
  Conference on Machine Learning (ICML)}, pp.~5025--5034, 2018.

\bibitem{li2019nattack}
Y.~Li, L.~Li, L.~Wang, T.~Zhang, and B.~Gong, ``Nattack: Learning the
  distributions of adversarial examples for an improved black-box attack on
  deep neural networks,'' in {\em International Conference on Machine Learning
  (ICML)}, pp.~3866--3876, 2019.

\bibitem{szegedy2017inception}
C.~Szegedy, S.~Ioffe, V.~Vanhoucke, and A.~A. Alemi, ``Inception-v4,
  inception-resnet and the impact of residual connections on learning,'' in
  {\em Proceedings of the Thirty-First AAAI Conference on Artificial
  Intelligence (AAAI)}, pp.~4278--4284, 2017.

\end{thebibliography}

%
\vspace{-1cm}
\begin{IEEEbiography}[{\includegraphics[width=1in,height=1.25in,clip,keepaspectratio]{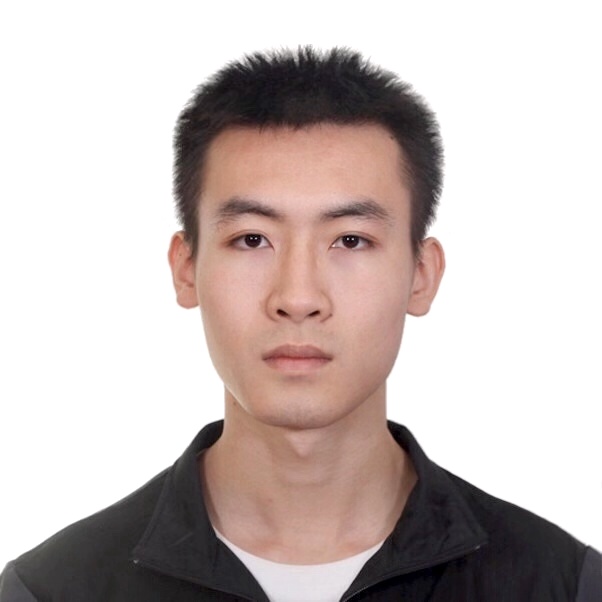}}]{Yinpeng Dong}
received his BS degree from the Department of Computer Science and Technology in Tsinghua University. He is currently a PhD student in the Department of Computer Science and Technology in Tsinghua University. His research interests are primarily on the adversarial robustness of machine learning and deep learning. He received Microsoft Research Asia Fellowship and Baidu Fellowship.
\end{IEEEbiography}\vspace{-1cm}

\begin{IEEEbiography}[{\includegraphics[width=1in,height=1.25in,clip,keepaspectratio]{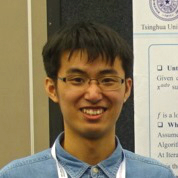}}]{Shuyu Cheng}
received his BS degree from the Department of Computer Science and Technology in Tsinghua University. He is currently a PhD student in the Department of Computer Science and Technology in Tsinghua University. His research interests are primarily on the adversarial robustness of machine learning and deep learning.
\end{IEEEbiography}\vspace{-1cm}

\begin{IEEEbiography}[{\includegraphics[width=1in,height=1.25in,clip,keepaspectratio]{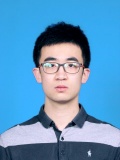}}]{Tianyu Pang} received his BS degree from the Department of Physics in Tsinghua University. He is currently a PhD student in the Department of Computer Science and Technology in Tsinghua University. His research interests are primarily on the adversarial robustness of machine learning and deep learning. He received Microsoft Research Asia Fellowship and Baidu Fellowship.
\end{IEEEbiography}\vspace{-1cm}

\begin{IEEEbiography}[{\includegraphics[width=1in,height=1.25in,clip,keepaspectratio]{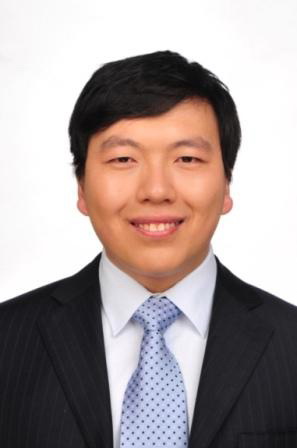}}]{Hang Su} is an associated professor in the Department of Computer Science and Technology at Tsinghua University. His research interests lie in the development of computer vision and machine learning algorithms for solving scientific and engineering problems arising from artificial learning and reasoning. He received ``Young Investigator Award'' from MICCAI2012, the ``Best Paper Award'' in AVSS2012, and ``Platinum Best Paper Award'' in ICME2018.
\end{IEEEbiography}\vspace{-1cm}

\begin{IEEEbiography}[{\includegraphics[width=1in,height=1.25in,clip,keepaspectratio]{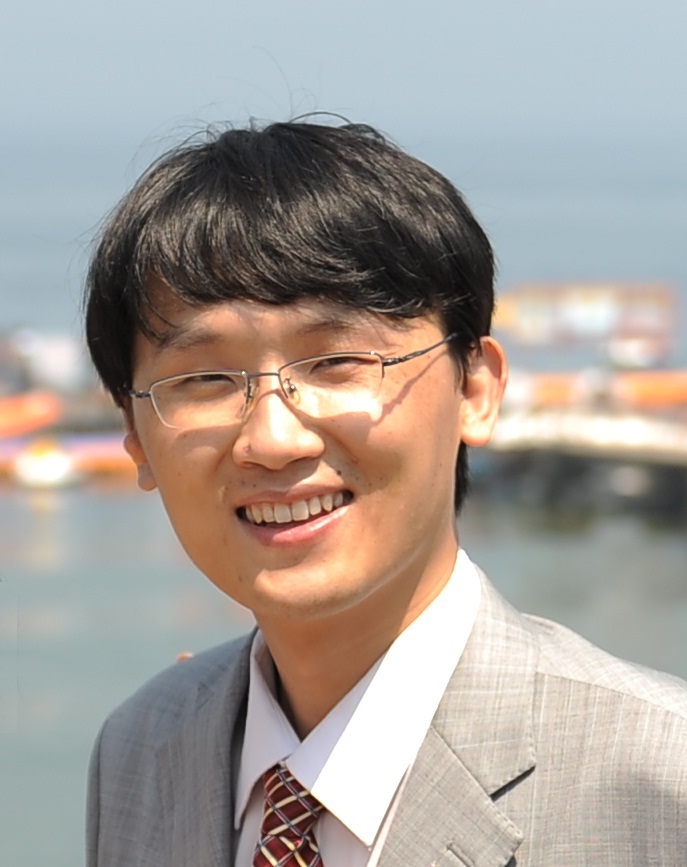}}]{Jun Zhu} received his BS and PhD degrees from the Department of Computer Science and Technology in Tsinghua University, where he is currently a professor. He was an adjunct faculty and postdoctoral fellow in the Machine Learning Department, Carnegie Mellon University. His research interest is primarily on developing machine learning methods to understand scientific and engineering data arising from various fields. He regularly serves as Area Chairs at prestigious conferences, including ICML, NeurIPS, ICLR, IJCAI and AAAI. He is a senior member of the IEEE, and was selected as ``AI's 10 to Watch'' by IEEE Intelligent Systems. 
\end{IEEEbiography}




\onecolumn
\normalsize
\setcounter{table}{3}
\setcounter{remark}{2}
\setcounter{algorithm}{2}
\numberwithin{equation}{section}
\setcounter{theorem}{0}

\appendices
\section{Proofs}
We provide the proofs in this section.
\subsection{Proof of Theorem 1}
\label{sec:proof_the-1}

\begin{theorem} If $f$ is differentiable at $x$, the loss of the gradient estimator $\hat{g}$ defined in Eq.~(5) is
\begin{equation*}
    \lim_{\sigma\to 0}L(\hat{g})=\|\nabla f(x)\|_2^2-\frac{\big(\nabla f(x)^\top\mathbf{C}\nabla f(x)\big)^2}{(1-\frac{1}{q})\nabla f(x)^\top\mathbf{C}^2\nabla f(x)+\frac{1}{q}\nabla f(x)^\top\mathbf{C}\nabla f(x)},
\end{equation*}
where $\sigma$ is the sampling variance, $\mathbf{C}=\E[u_iu_i^\top]$ with $u_i$ being the random vector, $\|u_i\|_2=1$, and $q$ is the number of random vectors as in Eq.~(5).
\end{theorem}

\begin{remark}
Rigorously speaking, we assume $\nabla f(x)^\top \mathbf{C} \nabla f(x)\neq 0$ in the statement of the theorem (and also in the proof), since when $\nabla f(x)^\top \mathbf{C} \nabla f(x)= 0$, both the numerator and the denominator of the fraction above are zero. When $\nabla f(x)^\top \mathbf{C} \nabla f(x)=0$, $u_i^\top \nabla f(x)=0$ holds almost surely, which implies that $L(\hat{g})=\|\nabla f(x)\|^2$ regardless of the value of $\sigma$. In fact, this case will not happen almost surely. In the setting of black-box attacks, we cannot even design a $\mathbf{C}$ with trace 1 such that $\nabla f(x)^\top \mathbf{C} \nabla f(x)= 0$ since $\nabla f(x)$ is unknown.
\end{remark}

\begin{proof}
First, we derive $L(\hat{g})$ based on the assumption that the single estimate $\hat{g}_i$ in Eq.~(5) is equal to $u_i^\top\nabla f(x) \cdot u_i$, which will hold when $f$ is locally linear.
\begin{lemma}
Assume that the single estimate $\hat{g}_i$ in Eq.~(5) is equal to $u_i^\top\nabla f(x) \cdot u_i$. We have
\begin{equation}
    L(\hat{g})=\|\nabla f(x)\|_2^2-\frac{(\nabla f(x)^\top\mathbf{C}\nabla f(x))^2}{(1-\frac{1}{q})\nabla f(x)^\top\mathbf{C}^2\nabla f(x)+\frac{1}{q}\nabla f(x)^\top\mathbf{C}\nabla f(x)}.
    \label{eq:biased_obj_app}
\end{equation}
\end{lemma}
\begin{proof}
First, we have \[\E\|\nabla f(x)-b\hat{g}\|_2^2=\|\nabla f(x)\|_2^2-2b\nabla f(x)^\top\E[\hat{g}]+b^2\E\|\hat{g}\|_2^2.\]
We have $\nabla f(x)^\top \E[\hat{g}]=\nabla f(x)^\top \E[\hat{g}_i]=\E[\nabla f(x)^\top u_i u_i^\top \nabla f(x)]=\E[(\nabla f(x)^\top u_i)^2]\geq 0$. Hence
\begin{equation}
    L(\hat{g})=\min_{b\geq0} \E\|\nabla f(x)-b\hat{g}\|_2^2 = \min_b \E\|\nabla f(x)-b\hat{g}\|_2^2 = \|\nabla f(x)\|_2^2-\frac{(\nabla f(x)^\top \E[\hat{g}])^2}{\E\|\hat{g}\|_2^2}.
    \label{eq:ghat_abstract}
\end{equation}
Since $\hat{g}_i=u_i^\top \nabla f(x) \cdot u_i$, and $u_i^\top u_i\equiv 1$, we have
\begin{align*}
    \E[\hat{g}_i] &= \mathbf{C}\nabla f(x), \\
    \E\|\hat{g}_i\|_2^2 &= \E[\hat{g}_i^\top \hat{g}_i] \\ &= \E[\nabla f(x)^\top u_i u_i^\top u_i u_i^\top \nabla f(x)] \\ &= \nabla f(x)^\top\E[u_i (u_i^\top u_i) u_i^\top]\nabla f(x) \\&= \nabla f(x)^\top\E[u_i  u_i^\top]\nabla f(x) \\&= \nabla f(x)^\top\mathbf{C}\nabla f(x).
\end{align*}
Given $\E[\hat{g}_i]$ and $\E\|\hat{g}_i\|^2$, the corresponding moments of $\hat{g}$ can be computed as
\begin{align}
    \E[\hat{g}]&=\E[\hat{g}_i]=\mathbf{C}\nabla f(x), \label{eq:ghat_as_gi}\\
    \E\|\hat{g}\|_2^2&=\E\|\hat{g}-\E[\hat{g}]\|_2^2+\|\E[\hat{g}]\|_2^2 \nonumber\\
    &=\frac{1}{q}\E\|\hat{g}_i-\E[\hat{g}_i]\|_2^2+\|\E[\hat{g}_i]\|_2^2 \nonumber\\
    &=\frac{1}{q}\E\|\hat{g}_i\|_2^2+(1-\frac{1}{q})\|\E[\hat{g}_i]\|_2^2 \label{eq:ghat_as_gi_2}\\
    &=(1-\frac{1}{q})\nabla f(x)^\top\mathbf{C}^2\nabla f(x)+\frac{1}{q}\nabla f(x)^\top\mathbf{C}\nabla f(x). \nonumber
\end{align}
Plug them into Eq.~\eqref{eq:ghat_abstract} and we complete the proof.
\end{proof}
Next, we prove that if $f$ is not locally linear, as long as it is differentiable at $x$, then by picking a sufficiently small $\sigma$, the loss tends to be that of the local linear approximation.
\begin{lemma}\label{thm:differentiable}
If $f$ is differentiable at $x$, let $L_0$ denote the right-hand side of Eq.~\eqref{eq:biased_obj_app}, then we have
\begin{equation*}
    \lim_{\sigma\to 0}L(\hat{g})=L_0.
\end{equation*}
\end{lemma}

\begin{proof}

Let $\hat{g}_i'=u_i^\top\nabla f(x)\cdot u_i$, $\hat{g}'=\frac{1}{q}\sum_{i=1}^q \hat{g}_i'$. Then $L_0=L(\hat{g}')$. By Eq.~\eqref{eq:ghat_abstract}, Eq.~\eqref{eq:ghat_as_gi}, and Eq.~\eqref{eq:ghat_as_gi_2}, it suffices to prove $\lim_{\sigma\to 0}\E[\hat{g}_i]=\E[\hat{g}_i']$ and $\lim_{\sigma\to 0}\E\|\hat{g}_i\|_2^2=\E\|\hat{g}_i'\|_2^2$.

For clarity, we redefine the notations. We omit the subscript $i$, make the dependence of $\hat{g}_i$ on $\sigma$ explicit (let $\hat{g}_\sigma$ denote $\hat{g}_i$), and let $\hat{g}_0$ denote $\hat{g}_i'$. Then we omit the hat in $\hat{g}$. That is, let  $g_0\triangleq u^\top \nabla f(x)\cdot u$ and $g_\sigma\triangleq \frac{f(x+\sigma u)-f(x)}{\sigma}\cdot u$, where $u$ is sampled uniformly from the unit hypersphere. Then we want to prove $\lim_{\sigma\to 0}\E[g_\sigma]=\E[g_0]$ and $\lim_{\sigma\to 0}\E\|g_\sigma\|_2^2=\E\|g_0\|_2^2$.

Since $f$ is differentiable at $x$, we have
\begin{equation}
    \lim_{\sigma\to 0}\sup_{\|u\|_2=1}\left|\frac{f(x+\sigma u)-f(x)}{\sigma}- u^\top\nabla f(x)\right|=0.
    \label{eq:differentiable}
\end{equation}
Since $\|u\|_2\equiv 1$, we have
\begin{align*}
    \lim_{\sigma\to 0}\E\|g_\sigma-g_0\|_2&\leq \lim_{\sigma\to 0}\sup_{\|u\|_2=1}\left|\frac{f(x+\sigma u)-f(x)}{\sigma}-u^\top\nabla f(x)\right|=0,\\
    \lim_{\sigma\to 0}\E\|g_\sigma-g_0\|_2^2&\leq \lim_{\sigma\to 0}\sup_{\|u\|_2=1}\left|\frac{f(x+\sigma u)-f(x)}{\sigma}-u^\top\nabla f(x)\right|^2=0.
\end{align*}
By applying Jensen's inequality to convex function $\|\cdot\|_2$, we have $\|\E[g_\sigma]-\E[g_0]\|_2\leq \E\|g_\sigma-g_0\|_2$. Since $\lim_{\sigma\to 0}\E\|g_\sigma-g_0\|_2=0$, and we have $\lim_{\sigma\to 0}\E[g_\sigma]=\E[g_0]$.

Since $\big|\|g_\sigma\|_2-\|g_0\|_2\big|\leq\|g_\sigma-g_0\|_2$, $\lim_{\sigma\to 0}\E\|g_\sigma-g_0\|_2=0$ and $\lim_{\sigma\to 0}\E\|g_\sigma-g_0\|_2^2=0$, we have $\lim_{\sigma\to 0}\E\big|\|g_\sigma\|_2-\|g_0\|_2\big|=0$ and $\lim_{\sigma\to 0}\E(\|g_\sigma\|_2-\|g_0\|_2)^2=0$. Also, we have $\|g_0\|_2\leq \|\nabla f(x)\|_2$. Hence, we have
\begin{align*}
    \lim_{\sigma\to 0}\big|\E\|g_\sigma\|_2^2-\E\|g_0\|_2^2\big|
    &\leq \lim_{\sigma\to 0}\E\big|\|g_\sigma\|_2^2-\|g_0\|_2^2\big| \\
    &=\lim_{\sigma\to 0}\E\Big[\big|\|g_\sigma\|_2-\|g_0\|_2\big|\big(\|g_\sigma\|_2+\|g_0\|_2\big)\Big] \\
    &\leq \lim_{\sigma\to 0}\E\Big[\big(\|g_\sigma\|_2-\|g_0\|_2\big)^2+2\|g_0\|_2\big|\|g_\sigma\|_2-\|g_0\|_2\big|\Big] \\
    &\leq \lim_{\sigma\to 0}\E\Big[\big(\|g_\sigma\|_2-\|g_0\|_2\big)^2+2\|\nabla f(x)\|_2\big|\|g_\sigma\|_2-\|g_0\|_2\big|\Big] \\
    &=0.
\end{align*}
The proof is complete.
\end{proof}
By combining the two lemmas above, our proof for Theorem \ref{the:1} is complete.
\end{proof}

\subsection{Proof of Eq.~(11)}
\label{sec:proof_11}
Suppose that $v$ is a fixed random vector and $\|v\|_2=1$. Let the $D$-dimensional random vector $u$ be
\begin{equation*}
    u=\sqrt{\lambda}\cdot v+\sqrt{1-\lambda}\cdot\overline{(\mathbf{I}-vv^\top)\xi},
\end{equation*}
where $\xi$ is sampled uniformly from the unit hypersphere. We need to prove that
\begin{equation*}
\mathbf{C}\equiv\E[u u^\top]=\lambda v v^\top + \frac{1-\lambda}{D-1}(\mathbf{I}-v v^\top).
\end{equation*}
\begin{proof}
Let $r\triangleq \overline{(\mathbf{I}-vv^\top)\xi}$. We choose an orthonormal basis $\{v_1,...,v_D\}$ of $\mathbb{R}^D$ such that $v_1=v$. Then $\xi$ can be written as $\xi=\sum_{i=1}^D a_i v_i$, where $a=(a_1,...,a_D)^\top$ is sampled uniformly from the unit hypersphere. Hence $(\mathbf{I}-vv^\top)\xi=\sum_{i=2}^D a_i v_i$, and $r=\frac{\sum_{i=2}^D a_i v_i}{\sqrt{\sum_{i=2}^D a_i^2}}$. Let $b_i=\frac{a_i}{\sqrt{\sum_{i=2}^D a_i^2}}$ for $i=2,3,...,D$, then $b=(b_2,b_3,...,b_D)^\top$ is sampled uniformly from the $(D-1)$-dimensional unit hypersphere, and $r=\sum_{i=2}^D b_i v_i$. Hence $\E[r]=0$. To compute $\E[r r^\top]$, we need a lemma first.
\begin{lemma}
Suppose that $d$ is a positive integer, $u=\sum_{i=1}^d a_i v_i$ where $a=(a_1,...,a_d)^\top$ is sampled uniformly from the $d$-dimensional unit hypersphere, then $\E[uu^\top]=\frac{1}{d}\sum_{i=1}^d v_i v_i^\top$.
\label{lem:covariance}
\end{lemma}
\begin{proof}
$\E[uu^\top] = \E[(\sum_{i=1}^d a_i v_i)(\sum_{j=1}^d a_j v_j^\top)] = \sum_{i=1}^d \sum_{j=1}^d v_i v_j^\top\E[a_i a_j]$. By symmetry, we have $\E[a_i a_j]=0$ when $i\neq j$, and $\E[a_i^2]=\E[a_j^2]$ for any $i, j$. Since $\sum_{i=1}^d a_i^2=1$, we have $\E[a_i^2]=\frac{1}{d}$ for any $i$. Hence $\E[uu^\top]=\frac{1}{d}\sum_{i=1}^d v_i v_i^\top$.
\end{proof}
Using the lemma, we have $\E[r r^\top]=\frac{1}{D-1}\sum_{i=2}^D v_i v_i^\top=\frac{1}{D-1}(\mathbf{I}-vv^\top)$. Since $\E[r]=0$, we have $\E[vr^\top]=\E[rv^\top]=0$. Hence, we have
\begin{align*}
    \E[uu^\top]&=\E[(\sqrt{\lambda}\cdot v+\sqrt{1-\lambda}\cdot r)(\sqrt{\lambda}\cdot v+\sqrt{1-\lambda}\cdot r)^\top] \\
    &=\lambda vv^\top+(1-\lambda)\E[rr^\top] \\
    &=\lambda vv^\top+\frac{1-\lambda}{D-1}(\mathbf{I}-vv^\top).
\end{align*}
The proof is complete.
\end{proof}
\begin{remark}
The construction of the random vector $u$ such that $\E[u u^\top]=\lambda v v^\top + \frac{1-\lambda}{D-1}(\mathbf{I}-v v^\top)$ is not unique. One can choose a different kind of distribution or simply take the negative of $u$ while remaining $\E[u u^\top]$ invariant.
\end{remark}

\subsection{Proof of Eq.~(12)}
\label{sec:proof_eq-lambda}
Let $\alpha=v^\top\overline{\nabla f(x)}$. Suppose that $D\geq 2$, $q\geq 1$. After plugging Eq.~(10) into Eq.~(9), the optimal $\lambda$ is given by
\begin{align}
    \lambda^* =
    \begin{cases}
        \hfil 0 & \text{if} \; \alpha^2\leq\dfrac{1}{D+2q-2}  \\
        \dfrac{(1-\alpha^2)(\alpha^2(D+2q-2)-1)}{2\alpha^2 Dq-\alpha^4 D(D+2q-2)-1} & \text{if} \; \dfrac{1}{D+2q-2} < \alpha^2 < \dfrac{2q-1}{D+2q-2} \\
        \hfil 1 & \text{if} \; \alpha^2 \geq \dfrac{2q-1}{D+2q-2}
    \end{cases}.
\end{align}

\begin{proof}
After plugging Eq.~(10) into Eq.~(9), we have
\begin{align*}
    L(\lambda)&=\|\nabla f(x)\|_2^2 \left(1-\frac{(\lambda \alpha^2+\frac{1-\lambda}{D-1}(1-\alpha^2))^2}{(1-\frac{1}{q})(\lambda^2 \alpha^2+(\frac{1-\lambda}{D-1})^2(1-\alpha^2))+\frac{1}{q}(\lambda \alpha^2+\frac{1-\lambda}{D-1}(1-\alpha^2))}\right).
\end{align*}
To minimize $L(\lambda)$, we should maximize
\begin{equation}
    F(\lambda)=\frac{(\lambda \alpha^2+\frac{1-\lambda}{D-1}(1-\alpha^2))^2}{(1-\frac{1}{q})(\lambda^2 \alpha^2+(\frac{1-\lambda}{D-1})^2(1-\alpha^2))+\frac{1}{q}(\lambda \alpha^2+\frac{1-\lambda}{D-1}(1-\alpha^2))}.
\label{eq:opt_F}
\end{equation}
Note that $F(\lambda)$ is a quadratic rational function w.r.t. $\lambda$.

Since we optimize $\lambda$ in a closed interval $[0, 1]$, checking $\lambda=0$, $\lambda=1$ and the stationary points (i.e., $F'(\lambda)=0$) would suffice. By solving $F'(\lambda)=0$, we have at most two solutions:
\begin{align}
    \lambda_1&=\frac{(1-\alpha^2)(\alpha^2(D+2q-2)-1)}{2\alpha^2 Dq-\alpha^4 D(D+2q-2)-1}, \label{eq:sol_lambda_1} \\
    \lambda_2&=\frac{1-\alpha^2}{1-\alpha^2 D} \nonumber,
\end{align}
where $\lambda_1$ or $\lambda_2$ is the solution if and only if the denominator is not 0. Given $\alpha^2\leq 1$ and $D\geq 2$, $\lambda_2\notin (0,1)$, so we only need to consider $\lambda_1$.

First, we figure out when $\lambda_1\in (0,1)$. We can verify that $\lambda_1=1$ when $\alpha^2=0$ and $\lambda_1=0$ when $\alpha^2=1$. Suppose that $\alpha^2\in (0,1)$. Let $J$ denote the numerator in Eq.~\eqref{eq:sol_lambda_1} and $K$ denote the denominator. We have that when $\alpha^2> \frac{1}{D+2q-2}$, $J> 0$; otherwise $J\leq 0$. We also have that when $\alpha^2< \frac{2q-1}{D+2q-2}$, $J < K$; otherwise $J\geq K$. Note that $J/K\in (0,1)$ if and only if $0< J < K$ or $0> J> K$. Hence, $\lambda_1\in (0,1)$ if and only if $\frac{1}{D+2q-2} < \alpha^2 < \frac{2q-1}{D+2q-2}$.

Case 1: $\lambda_1\notin (0,1)$. Then it suffices to compare $F(0)$ with $F(1)$. We have
\begin{equation*}
    F(0)=\frac{(1-\alpha^2)q}{D+q-2}, \; F(1)=\alpha^2.
\end{equation*}
Hence, $F(0)\geq F(1)$ if and only if $\alpha^2\leq \frac{q}{D+2q-2}$. It means that if $\alpha^2\geq\frac{2q-1}{D+2q-2}$, then $\lambda^*=1$; if $\alpha^2\leq\frac{1}{D+2q-2}$, then $\lambda^*=0$.

Case 2: $\lambda_1\in (0,1)$. After plugging Eq.~\eqref{eq:sol_lambda_1} into Eq.~\eqref{eq:opt_F}, we have
\begin{align}
    F(\lambda_1)&=\frac{4\alpha^2(1-\alpha^2)(q-1)q}{-1+2\alpha^2(D(2q-1)+2(q-1)^2)-\alpha^4(D+2q-2)^2}. \label{eq:F_lambda_1}
\end{align}
Now we prove that $F(\lambda_1)\geq F(0)$ and $F(\lambda_1)\geq F(1)$. Since when $0< \lambda < 1$, both the numerator and the denominator in Eq.~\eqref{eq:opt_F} is positive, we have $F(\lambda) > 0$, $\forall \lambda\in (0,1)$. Since the numerator in Eq.~\eqref{eq:F_lambda_1} is non-negative and $F(\lambda_1)>0$, we know that the denominator in Eq.~\eqref{eq:F_lambda_1} is positive. Hence, we have
\begin{align*}
    F(\lambda_1)-F(0)&=\frac{q(1-\alpha^2)(\alpha^2(D+2q-2)-1)^2}{(q+D-2)(-1+2\alpha^2(D(2q-1)+2(q-1)^2)-\alpha^4(D+2q-2)^2)}> 0; \\
    F(\lambda_1)-F(1)&=\frac{\alpha^2(\alpha^2(D+2q-2)+1-2q)^2}{-1+2\alpha^2(D(2q-1)+2(q-1)^2)-\alpha^4(D+2q-2)^2}> 0.
\end{align*}
Hence in this case $\lambda^*=\lambda_1$.

The proof is complete.
\end{proof}

\subsection{Monotonicity of $\lambda^*$}
We will prove that $\lambda^*$ is a monotonically increasing function of $\alpha^2$, and a monotonically decreasing function of $q$ (when $\alpha^2>\frac{1}{D}$).
\begin{proof}
To find the monotonicity w.r.t. $\alpha^2$, note that $\lambda^*=0$ if $\alpha^2\leq \frac{1}{D+2q-2}$ and $\lambda^*=1$ when $\alpha^2\geq \frac{2q-1}{D+2q-2}$. When $\frac{1}{D+2q-2}<\alpha^2<\frac{2q-1}{D+2q-2}$, we have
\begin{align}
    \lambda^* & = \dfrac{(1-\alpha^2)(\alpha^2(D+2q-2)-1)}{2\alpha^2 Dq-\alpha^4 D(D+2q-2)-1} \nonumber\\
    & = \dfrac{\alpha^4(D+2q-2)-\alpha^2(D+2q-1)+1}{\alpha^4D(D+2q-2)-2\alpha^2Dq+1} \nonumber\\
    & = \frac{1}{D}\left(1-\frac{(\alpha^2D-1)(D-1)}{\alpha^4D(D+2q-2)-2\alpha^2Dq+1}\right) \label{eq:lambda-mono}\\
    & = \frac{1}{D} - \frac{D-1}{\alpha^2D(D+2q-2) - (2Dq-D-2q+2) - 2\frac{(D-1)(q-1)}{\alpha^2D-1}}. \nonumber
\end{align}
When $\alpha^2 <\frac{1}{D}$ or $\alpha^2 >\frac{1}{D}$, a larger $\alpha^2$ leads to larger values of both $\alpha^2D(D+2q-2)$ and $-2\frac{(D-1)(q-1)}{\alpha^2D-1}$, and consequently leads to a larger $\lambda^*$. Meanwhile, by the argument in the proof of Eq.~(12), when $\frac{1}{D+2q-2}<\alpha^2<\frac{2q-1}{D+2q-2}$, the denominator of Eq.~\eqref{eq:sol_lambda_1} is positive, hence $\alpha^4D(D+2q-2)-2\alpha^2Dq+1<0$. By Eq.~\eqref{eq:lambda-mono}, when $\alpha^2 <\frac{1}{D}$, $\lambda^*<\frac{1}{D}$; when $\alpha^2 =\frac{1}{D}$, $\lambda^*=\frac{1}{D}$; when $\alpha^2 >\frac{1}{D}$, $\lambda^*>\frac{1}{D}$. We conclude that $\lambda^*$ is a monotonically increasing function of $\alpha^2$.

To find the monotonicity w.r.t. $q$ when $\alpha^2>\frac{1}{D}$, Eq.~\eqref{eq:lambda-1} tells us that when $q\leq\frac{\alpha^2(D-2)+1}{2(1-\alpha^2)}$, $\lambda^*=1$; else, $0<\lambda^*<1$. In the latter case, we rewrite Eq.~\eqref{eq:lambda-mono} as
\begin{align*}
    \lambda^* & = \frac{1}{D}\Big(1+\frac{(\alpha^2D-1)(D-1)}{2\alpha^2D(1-\alpha^2)q-\alpha^4D(D-2)-1}\Big).
\end{align*}
We have $(\alpha^2 D-1)(D-1)>0$, and as explained before, the denominator is positive for any $q$ such that $0<\lambda^*<1$. Hence, when $\alpha^2>\frac{1}{D}$, $\lambda^*$ is a monotonically decreasing function of $q$.
\end{proof}

\subsection{Proof of Theorem 2}
\begin{theorem}
    If $f$ is differentiable at $x$, the loss of the gradient estimator defined in Eq.~(13) is
    \begin{equation*}
        \lim_{\sigma\to 0} L(\hat{g})=\left(1-\frac{(\mu\alpha+(1-\mu)\E[\beta])^2}{\mu^2+(1-\mu)^2+2\mu(1-\mu)\alpha\E[\beta]}\right)\|\nabla f(x)\|_2^2,
    \end{equation*}
    where $\sigma$ is the sampling variance to get $\hat{g}^U$.
\end{theorem}

\begin{proof}
As in Eq.~(5), $\hat{g}^U=\frac{1}{q}\sum_{i=1}^q\hat{g}_i^U$ and $\hat{g}_i^U=\frac{f(x+\sigma u_i)-f(x)}{\sigma}\cdot u_i$, where $u_i$ is sampled from the uniform distribution on the $D$-dimensional unit hypersphere. First, we derive $L(\hat{g})$ based on the assumption that $\hat{g}_i^U$ is equal to $u_i^\top\nabla f(x) \cdot u_i$, which will hold when $f$ is locally linear.

\begin{lemma}
    Assume that $\hat{g}^U=\frac{1}{q}\sum_{i=1}^q (u_i^\top \nabla f(x)\cdot u_i)$ (then $\beta=\overline{\hat{g}^U}^\top\overline{\nabla f(x)}$). We have
    \begin{equation*}
        L(\hat{g})=\left(1-\frac{(\mu\alpha+(1-\mu)\E[\beta])^2}{\mu^2+(1-\mu)^2+2\mu(1-\mu)\alpha\E[\beta]}\right)\|\nabla f(x)\|_2^2.
    \end{equation*}
\end{lemma}

\begin{proof}
It can be verified\footnote{If $\hat{g}^U=0$, $\nabla f(x)^\top \hat{g}^U=\frac{1}{q}\sum_{i=1}^q (u_i^\top \nabla f(x))^2=0$, hence $u_i^\top\nabla f(x)=0$ for $i=1,2,...,q$, whose probability is 0. \label{fn:prob0}} that $\hat{g}^U=0$ happens with probability 0, hence we only consider $\hat{g}^U\neq 0$, which does not affect our conclusion. Then $\overline{\hat{g}^U}$ is always well-defined. The distribution of $\hat{g}^U$ is symmetric around the direction of $\nabla f(x)$, and so is the distribution of $\overline{\hat{g}^U}$. Hence we can suppose that $\E[\overline{\hat{g}^U}]=k\overline{\nabla f(x)}$. Since $\E[\beta]=\E[\overline{\hat{g}^U}]^\top\overline{\nabla f(x)}=k$, we have $\E[\overline{\hat{g}^U}]=\E[\beta]\overline{\nabla f(x)}$.

We have
\begin{align*}
    \E[\overline{\hat{g}^U}]^\top\nabla f(x) = \E[\beta]\overline{\nabla f(x)}^\top \nabla f(x) = \E[\beta]\|\nabla f(x)\|_2,
\end{align*}
and
\begin{align*}
    v^\top\E[\overline{\hat{g}^U}]=v^\top\E[\beta]\overline{\nabla f(x)}=\alpha\E[\beta].
\end{align*}
Together with $v^\top \nabla f(x)=\alpha\|\nabla f(x)\|_2$ and $\|v\|_2=1$, we have
\begin{align}
&\; \E\|\nabla f(x)-b\hat{g}\|_2^2 \nonumber \\ = & \; \E\|b\mu v + b(1-\mu)\overline{\hat{g}^U} - \nabla f(x)\|^2 \nonumber \\
= & \; b^2\mu^2 + b^2(1-\mu)^2 + \|\nabla f(x)\|^2_2 + 2b^2\mu(1-\mu)v^\top\E[ \overline{\hat{g}^U}] - 2b\mu\alpha\|\nabla f(x)\|_2 - 2b(1-\mu)\E[\overline{\hat{g}^U}]^\top\nabla f(x) \label{eq:average_proof} \\
= & \; b^2\mu^2 + b^2(1-\mu)^2 + \|\nabla f(x)\|^2_2 + 2b^2 \mu(1-\mu) \alpha \E[\beta] - 2b\mu\alpha\|\nabla f(x)\|_2 - 2b(1-\mu)\E[\beta]\|\nabla f(x)\| \nonumber \\ = & \; ((1-\mu)^2+\mu^2+2\mu(1-\mu)\alpha \E[\beta])b^2 -2(\alpha \mu+\E[\beta] (1-\mu))\|\nabla f(x)\|_2 b + \|\nabla f(x)\|^2_2. \nonumber
\end{align}
Since $\nabla f(x)^\top \hat{g}^U=\frac{1}{q}\sum_{i=1}^q (u_i^\top \nabla f(x))^2\geq 0$, then $\beta\geq 0$, and hence $\E[\beta]\geq 0$. Then $(1-\mu)^2+\mu^2+2\mu(1-\mu)\alpha \E[\beta]>0$ and $\alpha \mu+\E[\beta] (1-\mu)\geq 0$. Since $L(\hat{g})=\min_{b\geq 0}\E\|\nabla f(x)-b\hat{g}\|_2^2$, by optimizing the objective w.r.t. $b$ we complete the proof.
\end{proof}

Next, we prove that if $f$ is not locally linear, as long as it is differentiable at $x$, then by picking a sufficiently small $\sigma$, the loss tends to be that of the local linear approximation. Here, we redefine the notations as follows. We make the dependency of $\hat{g}^U$ on $\sigma$ explicit, i.e., we use $\hat{g}^U_\sigma$ to denote it. Meanwhile, we define $\hat{g}^U_0\triangleq \frac{1}{q}\sum_{i=1}^q (u_i^\top \nabla f(x)\cdot u_i)$ as the RGF estimator under the local linear approximation. We define $\hat{g}_\sigma=\mu v+(1-\mu)\overline{\hat{g}^U_\sigma}$ and $\hat{g}_0=\mu v+(1-\mu)\overline{\hat{g}^U_0}$. Then we have the following lemma.
\begin{lemma}
    If $f$ is differentiable at $x$, then
    \begin{equation*}
        \lim_{\sigma\to 0}L(\hat{g}_\sigma)=L(\hat{g}_0)
    \end{equation*}
\end{lemma}
\begin{proof}
By Eq.~\eqref{eq:average_proof}, it suffices to prove $\lim_{\sigma\to 0}\E[\overline{\hat{g}^U_\sigma]}=\E[\overline{\hat{g}^U_0}]$.

For any value of $u_1,u_2,...,u_q$, we have $\lim_{\sigma\to 0}\hat{g}^U_\sigma=\hat{g}^U_0$, i.e., $\hat{g}^U_\sigma$ converges pointwise to $\hat{g}^U_0$. Recall that $\mathrm{Pr}(\hat{g}^U_0=0)=0$, so we can only consider $\hat{g}^U_0\neq 0$, which does not affect our conclusion. Since $\overline{x}=\frac{x}{\|x\|_2}$ is continuous everywhere in its domain, $\overline{\hat{g}^U_\sigma}$ converges pointwise to $\overline{\hat{g}^U_0}$. Since the family $\{\overline{\hat{g}^U_\sigma}\}$ is uniformly bounded, by dominated convergence theorem we have $\lim_{\sigma\to 0}\E[\overline{\hat{g}^U_\sigma}]=\E[\overline{\hat{g}^U_0}]$.
\end{proof}
By combining the two lemmas above, our proof for Theorem~\ref{thm:average} is complete.
\end{proof}

\subsection{Proof of Eq.~(15)}
\label{sec:proof_eq-mu}
Let $\hat{g}$ be the PRGF-GA estimator with the balancing coefficient $\mu$ as defined in Eq.~(13). Let $L(\mu)=\lim_{\sigma\to 0} L(\hat{g})=\|\nabla f(x)\|_2^2-\frac{(\mu\alpha+(1-\mu)\E[\beta])^2}{\mu^2+(1-\mu)^2+2\mu(1-\mu)\alpha\E[\beta]}\|\nabla f(x)\|_2^2$. Then the optimal $\mu$ minimizing $L(\mu)$ is given by
\begin{align*}
    \mu^*=\frac{\alpha(1-\E[\beta]^2)}{\alpha(1-\E[\beta]^2) + (1-\alpha^2)\E[\beta]}.
\end{align*}
\begin{proof}
To minimize $L(\mu)$, we should maximize
\begin{equation*}
    F(\mu)=\frac{(\mu\alpha+(1-\mu)\E[\beta])^2}{\mu^2+(1-\mu)^2+2\mu(1-\mu)\alpha\E[\beta]}.
\end{equation*}
Note that $F(\mu)$ is a quadratic rational function w.r.t. $\mu$.

Since we optimize $\mu$ in a closed interval $[0, 1]$, checking $\mu=0$, $\mu=1$ and the stationary points (i.e. $F'(\mu)=0$) would suffice. By solving $F'(\mu)=0$, we have two solutions:
\begin{align*}
    \mu_1 &= \frac{\alpha(1-\E[\beta]^2)}{\alpha(1-\E[\beta]^2) + (1-\alpha^2)\E[\beta]}, \\
    \mu_2 &= \frac{\E[\beta]}{\E[\beta]-\alpha},
\end{align*}
where $\mu_2$ is the solution only when $\alpha\neq\beta$. Then we have
\begin{align*}
    F(0)&=\E[\beta]^2, \\
    F(1)&=\alpha^2, \\
    F(\mu_1)&=\frac{\alpha^2+\E[\beta]^2-2\alpha^2\E[\beta]^2}{1-\alpha^2\E[\beta]^2}, \\
    F(\mu_2)&=0.
\end{align*}
We have $F(0)\geq F(\mu_2)$, $F(1)\geq F(\mu_2)$, $F(\mu_1)-F(0)=\frac{\alpha^2(1-\E[\beta]^2)^2}{1-\alpha^2\E[\beta]^2}\geq 0$, $F(\mu_1)-F(1)=\frac{\E[\beta]^2(1-\alpha^2)^2}{1-\alpha^2\E[\beta]^2}\geq 0$. Therefore, the optimal solution of $\mu$ is $\mu^*=\mu_1$.
\end{proof}

\subsection{Proof of Eq.~(16)}\label{app:a7}
Let $\beta=\overline{\frac{1}{q}\sum_{i=1}^q (u_i^\top \nabla f(x)\cdot u_i)}^\top\overline{\nabla f(x)}$, we need to prove
\begin{equation*}
    \E[\beta] \approx \sqrt{\frac{q}{D+q-1}},
\end{equation*}
where $D$ and $q$ are the input dimension and the number of queries to get $\hat{g}^U$, respectively.
\begin{proof}
We let $\hat{g}_0^U=\frac{1}{q}\sum_{i=1}^q (u_i^\top \nabla f(x)\cdot u_i)$ as above. We can approximate $\E[\beta]$ by
\begin{align*}
    \E[\beta]&=\E[\sqrt{\beta^2}] \\&\approx \sqrt{\E[\beta^2]}\\&=\sqrt{1-\E[\min_b\|\overline{\nabla f(x)}-b\hat{g}_0^U\|^2]}\\&=\sqrt{1-\frac{1}{\|\nabla f(x)\|_2^2}\E[\min_b\|\overline{\nabla f(x)}-b\hat{g}_0^U\|^2]}\\&\approx \sqrt{1-\frac{1}{\|\nabla f(x)\|_2^2}\min_b\E\|\overline{\nabla f(x)}-b\hat{g}_0^U\|^2}\\&=\sqrt{1-\frac{1}{\|\nabla f(x)\|_2^2}L(\hat{g}_0^U)^2}.
\end{align*}
Here, the first equality is because that $\nabla f(x)^\top \hat{g}_0^U=\frac{1}{q}\sum_{i=1}^q (u_i^\top \nabla f(x))^2\geq 0$ and the second equality is because that we have $\min_b\|\overline{\nabla f(x)}-b\hat{g}_0^U\|^2=1-(\overline{\nabla f(x)}^\top \overline{\hat{g}_0^U})^2=1-\beta^2$. Intuitively, the two approximations work well because that the variances of $\beta$ and $\|\hat{g}_0^U\|_2$ are relatively small.

Now we define $F(\hat{g}_0^U)=1-\frac{1}{\|\nabla f(x)\|_2^2}L(\hat{g}_0^U)^2$. Then we have $\E[\beta]\approx \sqrt{F(\hat{g}_0^U)}$. Note that when $u_i$ is sampled from the uniform distribution on the unit hypersphere, $F(\hat{g}_0^U)$ is in fact $F(\frac{1}{D})$ in Eq.~\eqref{eq:opt_F}, since $\hat{g}_0^U$ is an RGF estimator w.r.t. locally linear $f$, and $\E[u_i u_i^\top]=\frac{1}{D}\mathbf{I}$ which corresponds to $\lambda=\frac{1}{D}$ in Eq.~(10). We can calculate $F(\frac{1}{D})=\frac{q}{D+q-1}$. Hence, $\E[\beta]\approx\sqrt{\frac{q}{D+q-1}}$.
\end{proof}

\subsection{Proof of Eq.~(22)}
Let $\alpha=v^\top\overline{\nabla f(x)}$, $A^2=\sum_{j=1}^d (v_j^\top\overline{\nabla f(x)})^2$. Suppose that $\alpha^2\leq 1$, $d\geq 1$, $q\geq 1$. After plugging Eq.~(21) into Eq.~(9), the optimal $\lambda$ is given by
\begin{align*}
\small
    \lambda^* =
    \begin{cases}
        \hfil 0 & \text{if} \; \alpha^2 \leq \dfrac{A^2}{d+2q-2} \\
        \dfrac{A^2(A^2-\alpha^2(d+2q-2))}{A^4+\alpha^4d^2-2A^2\alpha^2(q+dq-1)} & \text{if} \; \dfrac{A^2}{d+2q-2}<\alpha^2 < \dfrac{A^2(2q-1)}{d}\\
        \hfil 1 & \text{if} \; \alpha^2\geq \dfrac{A^2(2q-1)}{d}
    \end{cases}.
    \label{eq:lambda-2}
\end{align*}
\begin{proof}
The proof is very similar to that in Appendix~\ref{sec:proof_eq-lambda}. After plugging Eq.~(21) into Eq.~(9), we have
\begin{align*}
    L(\lambda)&=\|\nabla f(x)\|_2^2 \left(1-\frac{(\lambda \alpha^2+\frac{1-\lambda}{d}A^2)^2}{(1-\frac{1}{q})(\lambda^2 \alpha^2+(\frac{1-\lambda}{d})^2 A^2)+\frac{1}{q}(\lambda \alpha^2+\frac{1-\lambda}{d}A^2)}\right).
\end{align*}
To minimize $L(\lambda)$, we should maximize
\begin{equation}
    F(\lambda)=\frac{(\lambda \alpha^2+\frac{1-\lambda}{d}A^2)^2}{(1-\frac{1}{q})(\lambda^2 \alpha^2+(\frac{1-\lambda}{d})^2 A^2)+\frac{1}{q}(\lambda \alpha^2+\frac{1-\lambda}{d}A^2)}.
\label{eq:opt_F_dp}
\end{equation}
Note that $F(\lambda)$ is a quadratic rational function w.r.t. $\lambda$.

Since we optimize $\lambda$ in a closed interval $[0, 1]$, checking $\lambda=0$, $\lambda=1$ and the stationary points (i.e., $F'(\lambda)=0$) would suffice. By solving $F'(\lambda)=0$, we have at most two solutions:
\begin{align}
    \lambda_1&=\frac{A^2(\alpha^2(d+2q-2)-A^2)}{2A^2\alpha^2 (dq+q-1)-\alpha^4 d^2-A^4}, \label{eq:sol_lambda_1_dp} \\
    \lambda_2&=\frac{A^2}{A^2-\alpha^2 d} \nonumber,
\end{align}
where $\lambda_1$ or $\lambda_2$ is the solution if and only if the denominator is not 0. $\lambda_2\notin (0,1)$, so we only need to consider $\lambda_1$.

First, we figure out when $\lambda_1\in (0,1)$. We can verify that $\lambda_1=1$ when $\alpha^2=0$ and $\lambda_1=0$ when $A^2=0$. Suppose $\alpha^2\neq 0$ and $A^2\neq 0$. Let $J$ denote the numerator in Eq.~\eqref{eq:sol_lambda_1_dp} and $K$ denote the denominator. We have that when $\alpha^2> \frac{A^2}{d+2q-2}$, $J> 0$; otherwise $J\leq 0$. We also have that when $\alpha^2< \frac{A^2(2q-1)}{d}$, $J < K$; otherwise $J\geq K$. Note that $J/K\in (0,1)$ if and only if $0< J < K$ or $0> J> K$. Hence, $\lambda_1\in (0,1)$ if and only if $\frac{A^2}{d+2q-2} < \alpha^2 < \frac{A^2(2q-1)}{d}$.

Case 1: $\lambda_1\notin (0,1)$. Then it suffices to compare $F(0)$ and $F(1)$. We have
\begin{equation*}
    F(0)=\frac{A^2 q}{d+q-1}, F(1)=\alpha^2.
\end{equation*}
Hence, $F(0)\geq F(1)$ if and only if $\alpha^2\leq \frac{A^2 q}{d+q-1}$. It means that if $\alpha^2\geq\frac{A^2(2q-1)}{d}$, then $\lambda^*=1$; if $\alpha^2\leq\frac{A^2}{d+2q-2}$, then $\lambda^*=0$.

Case 2: $\lambda_1\in (0,1)$. After plugging Eq.~\eqref{eq:sol_lambda_1_dp} into Eq.~\eqref{eq:opt_F_dp}, we have
\begin{align}
    F(\lambda_1)&=\frac{4A^2\alpha^2 (A^2+\alpha^2)(q-1)q}{2A^2\alpha^2(2q(d+q-1)-d)-\alpha^4 d^2-A^4}. \label{eq:F_lambda_1_dp}
\end{align}

Now we prove that $F(\lambda_1)\geq F(0)$ and $F(\lambda_1)\geq F(1)$.
Since when $0<\lambda<1$, both the numerator and the denominator in Eq.~\eqref{eq:opt_F_dp} is positive, we have $F(\lambda)>0$, $\forall \lambda\in (0,1)$. Since the numerator in Eq.~\eqref{eq:F_lambda_1_dp} is non-negative, and $F(\lambda_1)> 0$, we know that the denominator in Eq.~\eqref{eq:F_lambda_1_dp} is positive. Hence, we have
\begin{align*}
    F(\lambda_1)-F(0)&=\frac{q A^2(\alpha^2(d+2q-2)-A^2)^2}{(q+d-1)(2A^2\alpha^2(2q(d+q-1)-d)-\alpha^4 d^2-A^4)}> 0; \\
    F(\lambda_1)-F(1)&=\frac{\alpha^2(\alpha^2 d+A^2(1-2q))^2}{2A^2\alpha^2(2q(d+q-1)-d)-\alpha^4 d^2-A^4}> 0.
\end{align*}
Hence in this case $\lambda^*=\lambda_1$.

The proof is complete.
\end{proof}

\subsection{Explanation on Eq.~(23)}
We explain why the construction of $u_i$ in Eq.~(23) makes $\E[u_iu_i^\top]$ a good approximation of $\mathbf{C}$.

Recall the setting: In $\mathbb{R}^D$, we have a normalized transfer gradient $v$, and a specified $d$-dimensional subspace with $\{v_1,...,v_d\}$ as its orthonormal basis. Let $\mathbf{C}=\lambda v v^\top+\frac{1-\lambda}{d}\sum_{j=1}^d v_j v_j^\top$. Here we argue that if $u=\sqrt{\lambda}\cdot v+\sqrt{1-\lambda}\cdot\overline{(\mathbf{I}-vv^\top)\mathbf{V}\xi}$, then $\E[u u^\top]\approx \mathbf{C}$.

Let $r\triangleq \overline{(\mathbf{I}-vv^\top)\mathbf{V}\xi}$. The reason why $\E[u u^\top]\neq \mathbf{C}$ is that $\E[rr^\top]\neq \frac{1}{d}\sum_{j=1}^d v_j v_j^\top$ when $v$ is not orthogonal to the subspace spanned by $\{v_1,...,v_d\}$. However, by symmetry, we still have $\E[r]=0$. To get an expression of $\E[rr^\top]$, we let $v_T$ denotes the projection of $v$ onto the subspace, and let $v_1=\overline{v_T}$ so that $v_2,...,v_d$ are orthonormal to $v_T$ (hence also orthonormal to $v$). We temporarily assume $v_T\neq v$ and $v_T\neq 0$. Now let $v_1'=\overline{(\mathbf{I}-vv^\top)v_T}=\overline{v_T-v^\top v_T\cdot v}$, then $\{v_1',v_2,...,v_d\}$ form an orthonormal basis of the subspace in which $r$ lies, and $v$ is orthogonal to this modified subspace. Now we have $\E[rr^\top]=\lambda_1 v_1'v_1'^\top + \frac{1-\lambda_1}{d-1}\sum_{j=2}^d v_j v_j^\top$ where $\lambda_1$ is a number in $[0, \frac{1}{d}]$. Note that when $v=v_T$, although $v_1'$ cannot be defined, we have $\lambda_1=0$. When $v_T=0$, we can just set $v_1'=v_1$ and $\lambda_1=\frac{1}{d}$. When $d$ is large, $\lambda_1$ is small, so for approximation we can replace $v_1'$ with $v_1$; $|\lambda-\frac{1}{d}|$ is small, so for approximation we can set $\lambda_1=\frac{1}{d}$. Then we have $\E[rr^\top]\approx \frac{1}{d}\sum_{j=1}^d v_j v_j^\top$. Since $\E[r]=0$, we have $\E[uu^\top]=\lambda vv^\top + (1-\lambda)\E[rr^\top]\approx \lambda vv^\top + \frac{1-\lambda}{d}\sum_{j=1}^d v_j v_j^\top$.

\begin{remark}
To avoid approximation, one can choose the subspace as spanned by $\{v_1',v_2,...,v_d\}$ instead of $\{v_1,v_2,...,v_d\}$ to ensure that $v$ is orthogonal to the subspace. Then $u$ can be sampled as
\begin{equation*}
    u=\sqrt{\lambda}\cdot v+\sqrt{1-\lambda}\cdot\overline{\mathbf{V}'\xi},
\end{equation*}
where $\mathbf{V}'=[v_1',v_2,...,v_d]$ and $\xi$ is sampled uniformly from the $d$-dimensional unit hypersphere. Note that here the optimal $\lambda$ is calculated using $A'^2=v_1'^\top\overline{\nabla f(x)}+\sum_{j=2}^d (v_j^\top\overline{\nabla f(x)})^2$. However, in practice, it is not convenient to make the subspace dependent on $v$, and the computational complexity is high to construct an orthonormal basis with one vector ($v_1'$) specified.
\end{remark}

\subsection{Proof of Theorem 3}
\begin{theorem}
Let $\alpha_1=v^\top\overline{\nabla f(x)}_T$. If $f$ is differentiable at $x$ and $A^2>0$, the loss of the gradient estimator define in Eq.~(24) is
\begin{equation*}
    \lim_{\sigma\to 0} L(\hat{g})= \left(1 -\frac{(\mu\alpha+(1-\mu)\E[\beta])^2}{\mu^2+(1-\mu)^2+2\mu(1-\mu)\frac{\alpha_1}{A^2}\E[\beta]}\right)\|\nabla f(x)\|^2,
\end{equation*}
where $\sigma$ is the sampling variance to get $\hat{g}^S$.
\end{theorem}

\begin{proof}
Similar to the proof of Theorem~\ref{thm:average}, we define $\hat{g}^S_0=\frac{1}{q}\sum_{i=1}^q (u_i^\top \nabla f(x)\cdot u_i)=\frac{1}{q}\sum_{i=1}^q (u_i^\top \nabla f(x)_T\cdot u_i)$, where $\nabla f(x)_T=\|\nabla f(x)\|_2\overline{\nabla f(x)}_T$ denotes the projection of $\nabla f(x)$ onto the subspace. 
Then $\beta=\overline{\hat{g}^S_0}^\top\overline{\nabla f(x)} = \overline{\hat{g}^S_0}^\top\overline{\nabla f(x)}_T$. Since $A^2>0$, we have $\nabla f(x)_T\neq 0$. As described in Footnote~\ref{fn:prob0}, we can prove $\mathrm{Pr}(\hat{g}^S_0=0)=0$ similarly. Now we only consider $\hat{g}^S_0\neq 0$. The distribution of $\hat{g}^S_0$ is symmetric around the direction of $\nabla f(x)_T$, and so is the distribution of $\overline{\hat{g}^S_0}$. Hence we can suppose that $\E[\overline{\hat{g}^S_0}]=k\overline{\nabla f(x)}_T$. Since $\E[\beta]=\E[\overline{\hat{g}^S_0}]^\top\overline{\nabla f(x)}_T =k\|\overline{\nabla f(x)}_T\|_2^2=kA^2$, we have $\E[\overline{\hat{g}_0^S}]=\frac{\E[\beta]}{A^2}\overline{\nabla f(x)}_T$.

Note that
\begin{align*}
    v^\top\E[\overline{\hat{g}_0^S}]=v^\top\frac{\E[\beta]}{A^2}\overline{\nabla f(x)}_T=\frac{\alpha_1}{A^2}\E[\beta].
\end{align*}
The rest of the proof is the same as that of Theorem~\ref{thm:average}.
\end{proof}

\subsection{Proof of Eq.~(26)}
Let $\hat{g}$ be the PRGF-GA estimator incorporating the data-dependent prior with the balancing coefficient $\mu$ as defined in Eq.~(24). Let $L(\mu)=\lim_{\sigma\to 0} L(\hat{g})=\|\nabla f(x)\|_2^2-\frac{(\mu\alpha+(1-\mu)\E[\beta])^2}{\mu^2+(1-\mu)^2+2\mu(1-\mu)\frac{\alpha_1}{A^2}\E[\beta]}\|\nabla f(x)\|_2^2$. Then the optimal $\mu$ minimizing $L(\mu)$ is given by
\begin{align*}
    \mu^*=\frac{A^2\alpha-\alpha_1\E[\beta]^2}{(A^2-\alpha_1\E[\beta])(\alpha+\E[\beta])}.
\end{align*}
\begin{proof}
The proof is very similar to that in Appendix~\ref{sec:proof_eq-mu}. To minimize $L(\mu)$, we should maximize
\begin{equation*}
    F(\mu)=\frac{(\mu\alpha+(1-\mu)\E[\beta])^2}{\mu^2+(1-\mu)^2+2\mu(1-\mu)\frac{\alpha_1}{A^2}\E[\beta]}.
\end{equation*}
Note that $F(\mu)$ is a quadratic rational function w.r.t. $\mu$.

Since we optimize $\mu$ in a closed interval $[0, 1]$, checking $\mu=0$, $\mu=1$ and the stationary points (i.e. $F'(\mu)=0$) would suffice. By solving $F'(\mu)=0$, we have two solutions:
\begin{align*}
    \mu_1 &= \frac{A^2\alpha-\alpha_1\E[\beta]^2}{(A^2-\alpha_1\E[\beta])(\alpha+\E[\beta])}, \\
    \mu_2 &= \frac{\E[\beta]}{\E[\beta]-\alpha},
\end{align*}
where $\mu_2$ is the solution only when $\alpha\neq\beta$. Then we have
\begin{align*}
    F(0)&=\E[\beta]^2, \\
    F(1)&=\alpha^2, \\
    F(\mu_1)&=\frac{A^4(\alpha^2+\E[\beta]^2)-2A^2\alpha\alpha_1\E[\beta]^2}{A^4-\alpha_1^2\E[\beta]^2}, \\
    F(\mu_2)&=0.
\end{align*}
We have $F(0)\geq F(\mu_2)$, $F(1)\geq F(\mu_2)$, $F(\mu_1)-F(0)=\frac{(A^2\alpha-\alpha_1\E[\beta]^2)^2}{A^4-\alpha_1^2\E[\beta]^2}\geq 0$, $F(\mu_1)-F(1)=\frac{\E[\beta]^2(A^2-\alpha\alpha_1)^2}{A^4-\alpha_1^2\E[\beta]^2}\geq 0$. Therefore, the optimal solution of $\mu$ is $\mu^*=\mu_1$.
\end{proof}

Let $\beta=\overline{\frac{1}{q}\sum_{i=1}^q (u_i^\top \nabla f(x)\cdot u_i)}^\top\overline{\nabla f(x)}$, in which $\{u_i\}_{i=1}^q$ lie in the subspace, we further need to prove
\begin{equation*}
    \E[\beta]\approx A\sqrt{\frac{q}{d+q-1}},
\end{equation*}
where $d$ is the subspace dimension, $q$ is the number of queries to get $\hat{g}^S$, and $A^2=\sum_{i=1}^d (v_i^\top \overline{\nabla f(x)})^2$.
\begin{proof}
Similar to the proof in Appendix~\ref{app:a7}, we approximate $\E[\beta]$ by $\sqrt{F(\hat{g}_0^S)}$, in which $F(\hat{g}_0^S)=1-\frac{1}{\|\nabla f(x)\|_2^2}L(\hat{g}_0^S)^2$, and $\hat{g}_0^S=\frac{1}{q}\sum_{i=1}^q (u_i^\top \nabla f(x)\cdot u_i)$. Note that when $u_i$ is sampled from the uniform distribution on the unit hypersphere in the subspace, $F(\hat{g}_0^S)$ is in fact $F(0)$ in Eq.~\eqref{eq:opt_F_dp}, since $\hat{g}_0^S$ is an RGF estimator w.r.t. locally linear $f$, and $\E[u_iu_i^T]=\frac{1}{d}\sum_{i=1}^d v_i v_i^\top$ which corresponds to $\lambda=0$ in Eq.~(21). We can calculate $F(0)=\frac{A^2q}{d+q-1}$. Hence, $\E[\beta]\approx A\sqrt{\frac{q}{d+q-1}}$.
\end{proof}

\section{Actual Implementation of PRGF-GA}
Note that in the PRGF-GA algorithm, the optimal coefficient $\mu^*$ in Eq.~(15) is calculated by minimizing the loss $L(\hat{g})$ of the gradient estimator defined as $\hat{g} = \mu v+(1-\mu) \overline{\hat{g}^U}$, where $v$ is the normalized transfer gradient and $\hat{g}^U$ is the ordinary RGF estimator. Since the loss $L(\hat{g})$ is a deterministic scalar whose computation requires taking expectation w.r.t. the randomness of $\hat{g}^U$, $\mu^*$ is a precomputed scalar which does not depend on the value of $\hat{g}^U$.
However, since $\mu$ is not concerned with the estimation process to get $\hat{g}^U$, 
we can actually obtain the value of $\hat{g}^U$ first and let $\mu$ depend on it, which could be beneficial when $\hat{g}^U$ exhibits high variance.

\begin{algorithm}[t]
\caption{Actual implementation of prior-guided random gradient-free algorithm based on gradient averaging (PRGF-GA)}
\label{alg:average-2}
\begin{algorithmic}[1]
\Require The black-box model $f$; input $x$ and label $y$; the normalized transfer gradient $v$; sampling variance $\sigma$; number of queries $q$; input dimension $D$; threshold $c$.
\Ensure Estimate of the gradient $\nabla f(x)$.
\State Estimate the cosine similarity $\alpha=v^\top\overline{\nabla f(x)}$ (detailed in Section~4.3);
\State Approximate $\E[\beta]$ by $\sqrt{\frac{q}{D+q-1}}$ as in Eq.~(16);
\State Calculate $\mu^*$ according to Eq.~(15) given $\alpha$ and $\E[\beta]$;
\If {$\mu^*\geq c$}
\Return $v$;
\EndIf
\State $\hat{g}^U \leftarrow \mathbf{0}$;
\For {$i = 1$ to $q$}
\State Sample $u_i$ from the uniform distribution on the $D$-dimensional unit hypersphere;
\State $\hat{g}^U \leftarrow \hat{g}^U + \dfrac{f(x + \sigma u_i,y) - f(x,y)}{\sigma} \cdot u_i$;
\EndFor
\State Estimate $v^\top \nabla f(x)$ by $\frac{f(x+\sigma v, y)-f(x,y)}{\sigma}$; Estimate $\overline{\hat{g}^U}^\top \nabla f(x)$ by $\frac{f(x+\sigma \overline{\hat{g}^U}, y)-f(x,y)}{\sigma}$;
\Return $\nabla f(x)\leftarrow v^\top\nabla f(x) \cdot v + \overline{\hat{g}^U} ^ \top \nabla f(x) \cdot \overline{\hat{g}^U}$.
\end{algorithmic}
\end{algorithm}

To this end, we need to calculate $\mu$ that leads to the best gradient estimator given the values of $v$ and $\hat{g}^U$.
We first assume that $v$ and $\overline{\hat{g}^U}$ are almost orthogonal with high probability, which is true in a high dimensional input space. (Without this assumption, we could perform Gram–Schmidt orthonormalization.) The problem is to find a vector in the subspace spanned by $v$ and $\overline{\hat{g}^U}$ that approximate the true gradient $\nabla f(x)$ best. This can be simply accomplished by projecting $\nabla f(x)$ onto the subspace, as
\begin{equation}
    \hat{g} = v^\top\nabla f(x) \cdot v + \overline{\hat{g}^U} ^ \top \nabla f(x) \cdot \overline{\hat{g}^U}.
\end{equation}
Therefore, the optimal $\mu$ can be expressed as
\begin{equation}
\begin{split}
    \mu^*=\frac{v^\top \nabla f(x)}{v^\top \nabla f(x) + \overline{\hat{g}^U}^\top \nabla f(x)}.
\end{split}
\end{equation}
$v^\top \nabla f(x)$ and $\overline{\hat{g}^U}^\top \nabla f(x)$ can be estimated by the finite difference method shown in Eq.~(17). We summarize the actual implementation of PRGF-GA in Algorithm~\ref{alg:average-2}.

\section{Estimation of $A$}
Suppose that the subspace is spanned by a set of orthonormal vectors $\{v_1,...,v_d\}$. Now we want to estimate 
\begin{align*}
    A^2=\sum_{j=1}^d (v_j^\top \overline{\nabla f(x)})^2=\frac{\sum_{j=1}^d (v_j^\top \nabla f(x))^2}{\|\nabla f(x)\|_2^2}=\frac{\|h(x)\|_2^2}{\|\nabla f(x)\|_2^2},
\end{align*}
where $h(x)=\sum_{j=1}^d v_j^\top\nabla f(x)\cdot v_j$ is the projection of $\nabla f(x)$ to the subspace. We can estimate $\|\nabla f(x)\|_2^2$ using the method introduced in Section~4.3. Here, we introduce the method to estimate $\|h(x)\|_2^2$.

Let $w=\mathbf{V}\xi$ where $\mathbf{V}=[v_1,v_2,...,v_d]$ and $\xi$ is a random vector uniformly sampled from the $d$-dimensional unit hypersphere. By Lemma~\ref{lem:covariance}, $\E[ww^\top]=\frac{1}{d}\sum_{j=1}^d v_j v_j^\top$. Suppose that we have $S$ i.i.d. such samples of $w$ denoted by $w_1, ..., w_S$, and we let $\mathbf{W}=[w_1, ..., w_S]$.

With $g(x_1, ..., x_S)=\frac{1}{S}\sum_{s=1}^S x_s^2$, we have
\begin{align*}
    g(\mathbf{W}^\top\nabla f(x))=g(\mathbf{W}^\top h(x))=\|h(x)\|_2^2\cdot g(\mathbf{W}^\top \overline{h(x)}).
\end{align*}
Hence $\frac{g(\mathbf{W}^\top\nabla f(x))}{\E[g(\mathbf{W}^\top \overline{h(x)})]}$ is an unbiased estimator of $\|h(x)\|_2^2$. Now, $\overline{h(x)}$ is in the subspace spanned by $\{v_1,...,v_d\}$, and $w_1$ is uniformly distributed on the unit hypersphere of this subspace. Hence $\E[(w_1^\top\overline{h(x)})^2]$ is independent of the direction of $\overline{h(x)}$ and can be computed. We have
\begin{align*}
    \E[g(\mathbf{W}^\top \overline{h(x)})]=\E[(w_1^\top\overline{h(x)})^2]=\overline{h(x)}^\top\E[w_1 w_1^\top]\overline{h(x)}=\overline{h(x)}^\top\frac{1}{d}\sum_{i=1}^d v_i v_i^\top\overline{h(x)}=\frac{1}{d}.
\end{align*}
Hence, we have the estimator $\|h(x)\|_2\approx \sqrt{\frac{d}{S}\sum_{s=1}^S(w_s^\top\nabla f(x))^2}$, where $w_s=\mathbf{V}\xi_s$ and $\xi_s$ is uniformly sampled from the unit hypersphere in $\mathbb{R}^d$. Finally we can get an estimate of $A$ by $A=\frac{\|h(x)\|_2}{\|\nabla f(x)\|_2}$.


\section{Additional Experiments}

\begin{table}
  \caption{The experimental results of black-box attacks against Inception-v3, VGG-16, and ResNet-50 under the $\ell_\infty$ norm on ImageNet. We report the attack success rate (ASR), and the average/median number of queries (AVG. Q/MED. Q) needed to generate an adversarial example over successful attacks. We mark the best results in \textbf{bold}.}
  \label{tab:final-results-linfty}
  \centering
  
  \begin{tabular}{l|ccc|ccc|ccc}
    \hline
    \multirow{2}{*}{Methods} & \multicolumn{3}{c|}{Inception-v3} & \multicolumn{3}{c|}{VGG-16} & \multicolumn{3}{c}{ResNet-50}\\
    \cline{2-10}
    & ASR & AVG. Q & MED. Q & ASR & AVG. Q & MED. Q & ASR & AVG. Q & MED. Q \\
    \hline
    NES~\cite{ilyas2018black} &  87.5\% & 1887 & 1122 & 95.6\% & 1507 & 1020 & 96.5\% & 1433 & 969 \\
    SPSA~\cite{uesato2018adversarial} & 93.6\% & 1766 & 1020 & 98.1\% & 1198 & 918 & 98.4\% & 1166 & 867 \\
    Bandits\textsubscript{T}~\cite{ilyas2018prior} & 89.5\% & 1891 & 952 & 93.8\% & 585 & 175 & 95.2\% & 1199 & 458\\
    Bandits\textsubscript{TD}~\cite{ilyas2018prior} & 94.7\% & 1099 & \bf330 & 95.1\% & \bf288 & \bf46 & 96.5\% & 651 & \bf158 \\
    $\mathcal{N}$ATTACK~\cite{li2019nattack} & \bf98.3\% & 1101 & 612 & \bf99.7\% & 639 & 408 & \bf99.5\% & 588 & 408 \\
    \hline
    RGF & 94.4\% & 1565 & 816 & 98.8\% & 1064 & 714 & \bf99.4\% & 990 & 663 \\
    PRGF-BS ($\lambda=0.05$) & 92.7\% & 1409 & 714 & 97.5\% & 1031 & 612 & 98.3\% & 891 & 561 \\
    PRGF-BS ($\lambda^*$) & 93.8\% & 979 & \bf414 & 98.5\% & 635 & 306 & 99.0\% & 507 & 236 \\
    PRGF-GA ($\mu=0.5$) & \bf94.9\% & 1263 & 624 & \bf98.9\% & 851 & 520 & 99.2\% & 758 & 468 \\
    PRGF-GA ($\mu^*$) & 94.8\% & \bf974 & 424 & 98.5\% & \bf560 & \bf298 & 99.3\% & \bf490 & \bf226 \\
    \hline
    RGF\textsubscript{D} & 97.2\% & 1034 & 561 & \bf100.0\% & 502 & 383 & 99.7\% & 595 & 408 \\
    PRGF-BS\textsubscript{D} ($\lambda=0.05$) & 97.7\% & 1005 & 510 & 99.9\% & 543 & 408 & 99.7\% & 598 & 408 \\
    PRGF-BS\textsubscript{D} ($\lambda^*$) & 97.3\% & 812 & 384 & 99.7\% & \bf370 & 262 & 99.6\% & 388 & \bf234 \\
    PRGF-GA\textsubscript{D} ($\mu=0.5$) & 98.0\% & 898 & 468 & \bf100.0\% & 481 & 364 & \bf99.8\% & 504 & 364 \\
    PRGF-GA\textsubscript{D} ($\mu^*$) & \bf98.4\% & \bf772 & \bf364 & 99.7\% & 374 & \bf246 & 99.6\% & \bf365 & 240 \\
    \hline
  \end{tabular}
\end{table}

\begin{table*}
  \caption{The experimental results of black-box attacks against ResNet-50, DenseNet-121, and SENet-18 under the $\ell_\infty$ norm on CIFAR-10. We report the attack success rate (ASR) and the average/median number of queries (AVG. Q/MED. Q) needed to generate an adversarial example over successful attacks. We mark the best results (including ASR $\geq$ 99.9\%) in \textbf{bold}.}
  \label{tab:cifar-linf}
  \centering
  \begin{tabular}{l|ccc|ccc|ccc}
    \hline
    \multirow{2}{*}{Methods} & \multicolumn{3}{c|}{ResNet-50} & \multicolumn{3}{c|}{DenseNet-121} & \multicolumn{3}{c}{SENet-18}\\
    \cline{2-10}
    & ASR & AVG. Q & MED. Q & ASR & AVG. Q & MED. Q & ASR & AVG. Q & MED. Q \\
    \hline
    NES~\cite{ilyas2018black} & 93.9\% & 781 & 408 & 96.1\% & 742 & 408 & 95.8\% & 699 & 357 \\
    SPSA~\cite{uesato2018adversarial} & \bf99.9\% & 627 & 408 & 99.8\% & 622 & 408 & \bf99.9\% & 571 & 357 \\
    Bandits\textsubscript{T}~\cite{ilyas2018prior} & \bf100.0\% & 372 & 186 & \bf100.0\% & 345 & 156 & \bf100.0\% & 312 & 142 \\
    $\mathcal{N}$ATTACK~\cite{li2019nattack} & \bf100.0\% & 384 & 255 & \bf100.0\% & 383 & 255 & \bf100.0\% & 343 & 204 \\
    \hline
    RGF & 98.4\% & 524 & 306 & 99.0\% & 499 & 306 & 99.1\% & 470 & 255 \\
    PRGF-BS ($\lambda=0.05$) & 99.2\% & 331 & 153 & 99.7\% & 275 & 153 & 99.7\% & 261 & 153 \\
    PRGF-BS ($\lambda^*$) & 99.6\% & 213 & 78 & \bf99.9\% & 206 & 113 & \bf99.9\% & 178 & 74 \\
    PRGF-GA ($\mu=0.5$) & 99.1\% & 310 & 153 & 99.7\% & 259 & 153 & 99.8\% & 229 & 153 \\
    PRGF-GA ($\mu^*$) & 99.6\% & \bf184 & \bf65 & \bf99.9\% & \bf156 & \bf65 & \bf99.9\% & \bf140 & \bf64 \\
    \hline
  \end{tabular}
\end{table*}

We show the experimental results of black-box adversarial attacks under the $\ell_\infty$ norm on ImageNet in Table~\ref{tab:final-results-linfty}, and on CIFAR-10 in Table~\ref{tab:cifar-linf}.

\end{document}